\documentclass[lettersize,journal]{IEEEtran}
\usepackage{amsmath,amsfonts}
\usepackage{algorithmic}
\usepackage{algorithm}
\usepackage{array}
\usepackage[caption=false,font=normalsize,labelfont=sf,textfont=sf]{subfig}
\usepackage{textcomp}
\usepackage{stfloats}
\usepackage{url}
\usepackage{xurl}
\usepackage{verbatim}
\usepackage{graphicx}
\usepackage{cite}
\usepackage{algorithm}
\usepackage{algorithmic}
\usepackage{booktabs}
\usepackage{adjustbox}
\usepackage{tabularx}
\usepackage{pifont}
\usepackage{amsmath}
\usepackage{multirow}
\usepackage{subcaption}
\usepackage{pdfpages}
\usepackage{placeins}
\usepackage{tcolorbox}      % for prompt boxes
\usepackage{listings}       % for code/json blocks
\usepackage{xcolor}         % for colors
\usepackage{multicol}  % 放在导言区
\usepackage{array}
\usepackage{caption}
\usepackage{hyperref}  % 确保导言区有
\usepackage{xspace}
\usepackage{makecell}
\lstdefinelanguage{json}{
    basicstyle=\ttfamily\footnotesize,
    numbers=none,
    breaklines=true,
    showstringspaces=false,
    columns=fullflexible,
    string=[b]",
    morestring=[b]',
    literate=
     *{0}{{{\color{blue}0}}}{1}
      {1}{{{\color{blue}1}}}{1}
      {2}{{{\color{blue}2}}}{1}
      {3}{{{\color{blue}3}}}{1}
      {4}{{{\color{blue}4}}}{1}
      {5}{{{\color{blue}5}}}{1}
      {6}{{{\color{blue}6}}}{1}
      {7}{{{\color{blue}7}}}{1}
      {8}{{{\color{blue}8}}}{1}
      {9}{{{\color{blue}9}}}{1}
      {:}{{{\color{red}:{}}}}{1}
      {,}{{{\color{red},{}}}}{1}
}
\usepackage{xcolor}
\definecolor{softred}{RGB}{255, 220, 220}
\definecolor{softgreen}{RGB}{220, 245, 220}

\newcommand{\hlred}[1]{\colorbox{softred}{#1}}
\newcommand{\hlgreen}[1]{\colorbox{softgreen}{#1}}
\begin{document}

%%
%% The "title" command has an optional parameter,
%% allowing the author to define a "short title" to be used in page headers.
\title{RW-Post: Auditable Evidence-Grounded Multimodal Fact-Checking in the Wild}

%%
%% The "author" command and its associated commands are used to define
%% the authors and their affiliations.
%% Of note is the shared affiliation of the first two authors, and the
%% "authornote" and "authornotemark" commands
%% used to denote shared contribution to the research.
\author{Danni Xu, Shaojing Fan, Harry Cheng, 
Mohan Kankanhalli, \IEEEmembership{Fellow, IEEE}
\thanks{
Author Danni Xu, Harry Cheng, and Mohan Kankanhalli are with the School of Computing (SoC),
National University of Singapore (NUS), Singapore.
Author Shaojing Fan is with the Department of Electrical and Computer Engineering (ECE), National University of Singapore (NUS), Singapore.
}}

\maketitle
\begin{abstract}
Multimodal misinformation increasingly leverages visual persuasion, where repurposed or manipulated images strengthen misleading text. 
We introduce \textbf{RW-Post}, a post-aligned \textbf{text--image benchmark} for real-world multimodal fact-checking with \emph{auditable} annotations: each instance links the original social-media post with reasoning traces and explicitly linked evidence items derived from human fact-check articles via an LLM-assisted extraction-and-auditing pipeline.
RW-Post supports controlled evaluation across closed-book, evidence-bounded, and open-web regimes, enabling systematic diagnosis of visual grounding and evidence utilization.
We provide \textbf{AgentFact} as a reference verification baseline and benchmark strong open-source LVLMs under unified protocols.
Experiments show substantial headroom: current models struggle with faithful evidence grounding, while evidence-bounded evaluation improves both accuracy and faithfulness.
Code and dataset will be released at 
\url{https://github.com/xudanni0927/AgentFact}.
\end{abstract}
\section{Introduction}
\label{sec:intro}

The proliferation of social media has accelerated the spread of persuasive multimodal misinformation, where images and videos are used to shape perception alongside misleading text, posing risks to public health and social behavior~\cite{borges2022infodemics, wsj2025falseTariffHeadline, ARCURI2023106130}. 
Recent advances in generative AI further lower the cost of producing realistic deceptive content~\cite{berman2024aimisinformation, DeepfakeSurvey2024Wang, reuters2025duterteDisinformation, 11010889}. 
These challenges motivate the need for automated systems capable of verifying multimodal claims with reliable evidence.

Existing approaches to multimodal misinformation detection often focus on narrow subtasks, such as out-of-context image detection or deepfake detection~\cite{qi2024sniffer, NoiseDF2023Wang, 10487975, 10290956, 9408664}. 
More recent work leverages LLMs and LVLMs with chain-of-thought prompting, claim decomposition, or retrieval augmentation for fact-checking~\cite{PanQACheck23, leite2023detecting, zhang2023towards, kareem2023fighting, xuan2024lemma, braun2024defame}. 
However, these systems frequently rely on shallow evidence usage and lack faithful grounding in verifiable sources. 
Progress is further limited by the absence of benchmarks that jointly provide real-world multimodal claims, auditable reasoning traces, and explicitly linked evidence.

To address this gap, we introduce \textbf{RW-Post}, a \textbf{text--image benchmark} for real-world multimodal fact-checking. 
RW-Post pairs fact-checked claims with their original social-media posts, preserving contextual cues critical for verification. 
The dataset additionally provides reasoning traces and explicitly linked evidence extracted from human-written fact-checking articles through an LLM-assisted extraction-and-auditing pipeline. 
RW-Post supports both \textbf{closed-book} verification using only the post context, \textbf{evidence-bounded} verification using oracle evidence in dataset, and \textbf{open-web} verification with external evidence retrieval. 
To ensure fair evaluation in the open-web setting, we standardize the search backend and apply strict anti-leakage filtering.

To establish reference baselines, we introduce \textbf{AgentFact}, a modular verification agent-based framework that integrates strategy planning, evidence retrieval, visual analysis, reasoning, and evidence-grounded explanation generation, combined by retrieval-reasoning iterated workflow.
We benchmark representative open-source LVLMs under unified prompting protocols and controlled evaluation settings. 
Experiments demonstrate that RW-Post reveals substantial headroom for multimodal verification: even strong baselines struggle with faithful evidence grounding and visually informed reasoning, highlighting promising directions for future research.

Our contributions are:
\begin{enumerate}
\item We introduce \textbf{RW-Post}, a benchmark for explainable multimodal fact-checking with post-aligned claims, reasoning traces, and source-grounded evidence.
\item We propose an LLM-assisted pipeline for extracting and auditing reasoning and evidence from human-written fact-checking articles.
\item We provide \textbf{AgentFact} and benchmark strong LVLMs under controlled closed-book, evidence-bounded, and open-web evaluation protocols.
\end{enumerate}
\section{Related Work}
\label{sec:related_works}
\subsection{Misinformation and OOC Detection Datasets}
Misinformation datasets can be broadly grouped into three categories: (1) \textit{collected datasets}, (2) \textit{synthetic datasets}, and (3) \textit{OOC-specific datasets}, as summarized in Table~\ref{tab:dataset_list_previous}.

\paragraph{Collected datasets.}
Early datasets compile verified claims from fact-checking websites such as PolitiFact and Snopes (e.g., FakeNewsNet~\cite{shu2020fakenewsnet}, LIAR-RAW~\cite{yang-etal-2022-coarse}, MuMiN~\cite{NielsenMcConville2022}). 
Many are text-only and suffer from noisy annotations or missing context, since social-media posts are often matched to claims via keyword or similarity heuristics. 
Large-scale corpora such as Fakeddit~\cite{nakamura2020r} further rely on weak labels from platform tags. 
AVERITEC~\cite{schlichtkrull2023averitec} improves annotation quality by extracting explanations from fact-checking articles, but remains limited to text-only claims.

Recent efforts introduce multimodal resources. 
CLAIMREVIEW2024+ manually links images to fact-checking claims, while Mocheg heuristically extracts rationales from Snopes articles. 
However, these datasets often lack original posts, reliable evidence links, or consistent explanations.

\paragraph{Synthetic and OOC datasets.}
Synthetic datasets increase scale by generating claims from structured sources such as Wikipedia (e.g., FEVER, FEVEROUS, HOVER) or LLM-generated content. 
However, they mainly focus on textual entailment and rarely capture multimodal interactions.
OOC datasets instead examine mismatches between real images and misleading text, using random mismatches (e.g., MAIM, COSMOS~\cite{jaiswal2017multimedia, aneja2023cosmos}), semantic mismatches (e.g., NewsCLIPpings~\cite{luo2021newsclippings}), or entity substitution~\cite{papadopoulos2023synthetic}. 
Despite these advances, synthetic and OOC datasets often lack realistic post context and interpretable explanations.

These limitations highlight the need for a realistic multimodal fact-checking benchmark that preserves \emph{post-level context} and provides \emph{auditable evidence}. 
\textbf{RW-Post} addresses this gap by aligning fact-checked claims with their original social-media posts and linking them to reasoning traces and verifiable evidence.

\subsection{Misinformation and OOC Detection Methods}

Existing approaches can be categorized along two axes: 
(1) internal estimation vs.\ external retrieval, and 
(2) feature-based vs.\ LLM/LVLM-based reasoning.

\textbf{Feature-based internal} methods rely on linguistic style, propagation patterns, or image–text consistency signals~\cite{przybyla2020capturing, shu2020hierarchical, zhou2020multimodal, luo2021newsclippings}. 
However, these approaches lack external grounding and often struggle with real-world variability.

\textbf{Feature-based external} approaches retrieve related captions or images and compare them using semantic or entity-level similarity~\cite{aneja2023cosmos, muller2020multimodal, abdelnabi2022open}. 
While retrieval improves robustness, these methods typically rely on shallow matching.

Recent \textbf{LLM/LVLM-based} systems leverage prompting and retrieval augmentation, sometimes in agent-style pipelines~\cite{PanQACheck23, xuan2024lemma, braun2024defame}, improving interpretability but still exhibiting superficial evidence use in complex multimodal verification.

A central challenge is \textbf{faithfulness}: models can generate fluent rationales while hallucinating reasoning or misusing evidence, where explanations are not supported by verifiable sources.
This has motivated evidence-grounded evaluation protocols that require citing evidence and auditing whether citations truly justify predictions, especially in open-web settings where retrieval adds variability and leakage risks.
These limitations highlight the need for benchmarks that provide explicitly linked evidence and reasoning traces and support controlled regimes (e.g., closed-book vs.\ evidence-bounded vs.\ open-web) for systematic auditing and diagnosis.

\begin{table}[ht]
\small
\centering
\caption{
Fact-checking and out-of-context (OOC) datasets.
$I$: image;
$C$: claim; $P$: post; $R$: reasoning traces; 
$E_{src}$: external evidence sources; $E_{link}$: reasoning--evidence alignment for auditing. 
\ding{51}/\ding{55} denote provided/not provided.
}
\label{tab:dataset_list_previous}
\setlength{\tabcolsep}{4pt}
\begin{tabularx}{\linewidth}{
l c
c c c c c
p{0.7cm}
}
\toprule
Dataset & $I$ & $C$ & $P$ & $R$ & $E_{src}$ & $E_{link}$ & Size \\
\midrule
\multicolumn{8}{l}{\textit{Synthetic Fact-checking Datasets}} \\
\midrule
HOVER~\cite{jiang2020hover} & \ding{55} & \ding{51} & \ding{55} & \ding{55} & \ding{51} &  \ding{55}   & 26k \\
FEVEROUS~\cite{aly2021fact} & \ding{55} & \ding{51} & \ding{55} & \ding{55} & \ding{51} &  \ding{55}   & 87k \\
Multi-News FC~\cite{chen2024metasumperceiver} & \ding{51} & \ding{51} & \ding{55} & \ding{55} & \ding{51} &  \ding{55}& 1291k \\
\midrule
\multicolumn{8}{l}{\textit{OOC-specific Datasets}} \\
\midrule
MMFakeBench~\cite{liu2024mmfakebench} & \ding{51} & \ding{55} & \ding{51} & \ding{55} & \ding{55} &  \ding{55}   & 11k \\
MAIM~\cite{jaiswal2017multimedia} & \ding{51} & \ding{55} & \ding{51} & \ding{55} & \ding{55} &  \ding{55}   & 239k\\
COSMOS~\cite{aneja2023cosmos} & \ding{51} & \ding{55} & \ding{51} & \ding{55} & \ding{55} &  \ding{55}   & 453k\\
NewsCLIPpings~\cite{luo2021newsclippings}  & \ding{51} & \ding{55} & \ding{51} & \ding{55} & \ding{55} &  \ding{55}   & 988k \\ 
\midrule
\multicolumn{8}{l}{\textit{Collected Fact-checking Datasets}} \\
\midrule
Twitter~\cite{weibo_twitter}  & \ding{51} & \ding{55} & \ding{51} & \ding{55} & \ding{55} & \ding{55} & 13k+ \\
Weibo~\cite{weibo_twitter} & \ding{51} & \ding{55} & \ding{51} & \ding{55} & \ding{55} &  \ding{55}  & 9k+ \\
FakeNewsNet~\cite{shu2020fakenewsnet} & \ding{51} & \ding{51} & \ding{51} & \ding{55} & \ding{55} &  \ding{55}  & 19k+ \\
Fakeddit~\cite{nakamura2020r}  & \ding{51} & \ding{55} & \ding{51} & \ding{55} & \ding{55} &  \ding{55}  & 1M+ \\
Weibo21~\cite{nan2021mdfend} & \ding{51} & \ding{55} & \ding{51} & \ding{55} & \ding{55} &  \ding{55}  & 9k+ \\
MuMiN~\cite{NielsenMcConville2022} & \ding{51} & \ding{51} & \ding{51} & \ding{55} & \ding{55} &  \ding{55}   & 21M+ \\
RAWFC~\cite{yang-etal-2022-coarse} & \ding{55} & \ding{51} & \ding{55} & \ding{51} & \ding{51} &  \ding{55} & 2k \\
LIAR-RAW~\cite{yang-etal-2022-coarse} & \ding{55} & \ding{51} & \ding{55} & \ding{51} & \ding{51} &  \ding{55}  & 12k \\
AVERITEC~\cite{schlichtkrull2023averitec} & \ding{55} & \ding{51} & \ding{51} & \ding{51} & \ding{51} &  \ding{55}  & 4.57k \\
Mocheg~\cite{mocheg23} & \ding{51} & \ding{51} & \ding{55} & \ding{51} & \ding{51} &  \ding{55}   & 15k \\
Factify~2~\cite{mishra2022factify} & \ding{51} & \ding{51} & \ding{51} & \ding{55} & \ding{51} &  \ding{55}  & 50k \\
MR$^{2}$~\cite{hu2023mr2} & \ding{51} & \ding{51} & \ding{51} & \ding{55} & \ding{51} &  \ding{55} & 14k+ \\
CLAIMREVIEW+~\cite{braun2024defame} & \ding{51} & \ding{51} & \ding{55} & \ding{55} & \ding{55} &  \ding{55}  & 0.3k \\
\textbf{RW-Post (ours)} & \ding{51} & \ding{51} & \ding{51} & \ding{51} & \ding{51} &  \ding{51} & 1.77k \\

\bottomrule
\end{tabularx}
\end{table}
\section{RW-Post: A Post-aligned Multimodal Fact-Checking Benchmark with Auditable Evidence}
\label{sec:dataset}
We introduce RW-Post, a benchmark for explainable multimodal fact-checking that reflects real-world verification scenarios.
Each claim is paired with its original social-media post, structured reasoning (logic and key points), and explicitly linked evidence.
Instead of using loosely relevant documents, RW-Post emphasizes targeted evidence that directly supports factual reasoning, aligning with the rigor of professional fact-checking. 

\textbf{Benchmark Regimes.}
RW-Post supports three complementary evaluation regimes:
(1) \textbf{closed-book}, where models observe only the post and image;
(2) \textbf{evidence-bounded}, where models receive the post together with a fixed set of evidence items (with stable IDs and source URLs); and
(3) \textbf{open-web}, where models retrieve evidence via web search under a standardized, leakage-controlled protocol.
These regimes enable systematic analysis of visual grounding, evidence utilization, and explanation faithfulness.

Following common benchmark practice~\cite{yu2024mm, hendryckstest2021, liu2024mmfakebench}, RW-Post is released with a development and a test split (1:5), comprising 252 and 1,268 samples, respectively. We use the dev split for prompt and system development and report all benchmark results on the test split; while our main experiments are zero-shot, the dataset can also support supervised finetuning.

\begin{figure*}
    \centering
    \includegraphics[width=\linewidth]{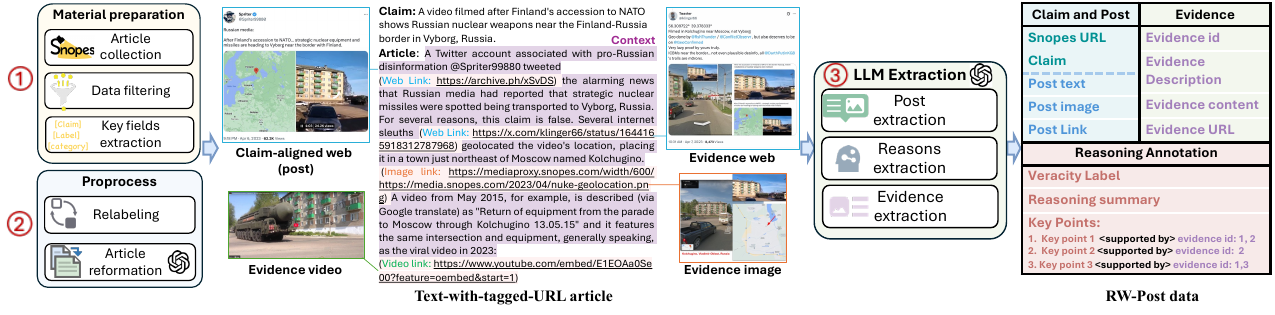}
    \caption{RW-Post Dataset: Use Context (purple highlight) helps LLM determine whether the link (pink highlight) is post or evidence and which rationale it supports.}
    \label{fig:rw_post_dataset_construction}
\end{figure*}

We collect 13,000 Snopes fact-checking articles (2017–2024) and apply strict filtering and quality control.
To enable structured extraction, webpages are reformatted into a text–with-tagged-URL representation that preserves the contextual position of each link.
This representation allows LLMs to reliably extract posts, images, reasoning, and evidence.
The pipeline is illustrated in Fig.~\ref{fig:rw_post_dataset_construction}.

\textbf{Data collection.}
We collect fact-checking articles from Snopes, which covers diverse misinformation topics and follows professional verification standards.
For each article, we extract the \textit{URL, category, headline, claim, truthfulness label, and full HTML content}.

\textbf{Text and image filtering.}
Articles shorter than 90 words or longer than 2,000 words are removed to avoid insufficient or overly verbose content.
Images smaller than 200×200 pixels are also excluded.

\subsection{Dataset annotation}
\textbf{Relabeling truthfulness labels.}
Snopes provides over ten fine-grained truthfulness labels, some of which have overlapping meanings (e.g., \textit{MISCAPTION} vs.\ \textit{FALSE}) and highly imbalanced distributions.
To simplify evaluation and improve label consistency, we consolidate them into five high-level categories based on factual accuracy:
\textit{True}, \textit{False}, \textit{Unproven}, \textit{Mixture}, and \textit{Outdated}.
Labels unrelated to factual correctness (e.g., satire) are excluded.
Table~\ref{tab:label-mapping} shows the mapping from original Snopes labels to our consolidated categories.

\begin{table}[ht]
\centering
\small
\begin{tabular}{lp{6.5cm}} 
\toprule
\textbf{Label} & \textbf{Original Labels} \\
\midrule
False & FALSE, MISCAPTION, MISCAPTIONED, FAKE, LEGEND, SCAM, MISATTRIBUTED \\

Mixture & MOSTLY FALSE, MIXTURE, MOSTLY TRUE \\

Outdated & OUTDATED \\

True & TRUE, CORRECT ATTRIBUTION, LEGIT \\

Unproven & UNFOUNDED, UNPROVEN, RESEARCH IN PROGRESS \\
\bottomrule
\end{tabular}
\caption{Mapping from original labels to high-level labels.}
\label{tab:label-mapping}
\end{table}

\textbf{Article reformatting.}
Inspired by Flamingo~\cite{flamingo-2022}, to enable structured extraction, we convert each HTML article into a \emph{text-with-tagged-URLs} representation. 
All URLs in the webpage are identified and categorized as image, video, or webpage links based on file extensions, and then inserted back into the text with modality tags while preserving their original positions. 
This representation maintains the contextual relationship between text and links, allowing LLMs to reliably identify claim-aligned posts, images, and evidence, and aligning evidence with reasoning.
Examples of this format are shown in Fig.~\ref{fig:rw_post_dataset_construction}.

\textbf{Explanation extraction.}
We design an LLM-assisted extraction pipeline to generate structured annotations from the reformatted articles.
Using GPT-4o, we extract:
(1) the original post and associated images,
(2) a concise reasoning overview and itemized key points, and
(3) evidence aligned to each key point with stable evidence IDs and URLs.

\subsection{Quality Control}

To ensure dataset quality, we apply both automated filtering and human verification.

\textbf{Multimodal relevance filtering.}
We remove claims where the image does not play a meaningful role in verification.
Specifically, GPT-based prompts determine whether the image serves as direct evidence, strengthens the claim’s credibility, or significantly affects its perceived truthfulness.
Claims judged as not requiring image context are excluded.

\textbf{Watermark leakage filtering.}
Images containing fact-checking watermarks (e.g., “fake”, “false”, or “misleading”) are removed to prevent label leakage.

Both steps combine GPT-based filtering with manual inspection.
Manual checks show high consistency with the automated filtering results, indicating reliable identification of multimodal claims and watermark-free images.
Prompt templates are provided in Appendix.

\subsection{Human validation}
To verify the reliability of the LLM-extracted annotations, we conduct a human validation study on posts, images, reasoning, and evidence.

\paragraph{Post and Image Validation}

Annotators verify whether extracted posts and images correctly correspond to the original fact-checking articles.
Each sample is independently annotated by two reviewers, with disagreements resolved through discussion.
On a manually reviewed set of 100 articles, the system correctly extracts post links in 76.4\% of link-containing cases, with most errors arising from missed extractions rather than incorrect matches.

\begin{figure}[t]
    \centering
    \includegraphics[width=\linewidth]{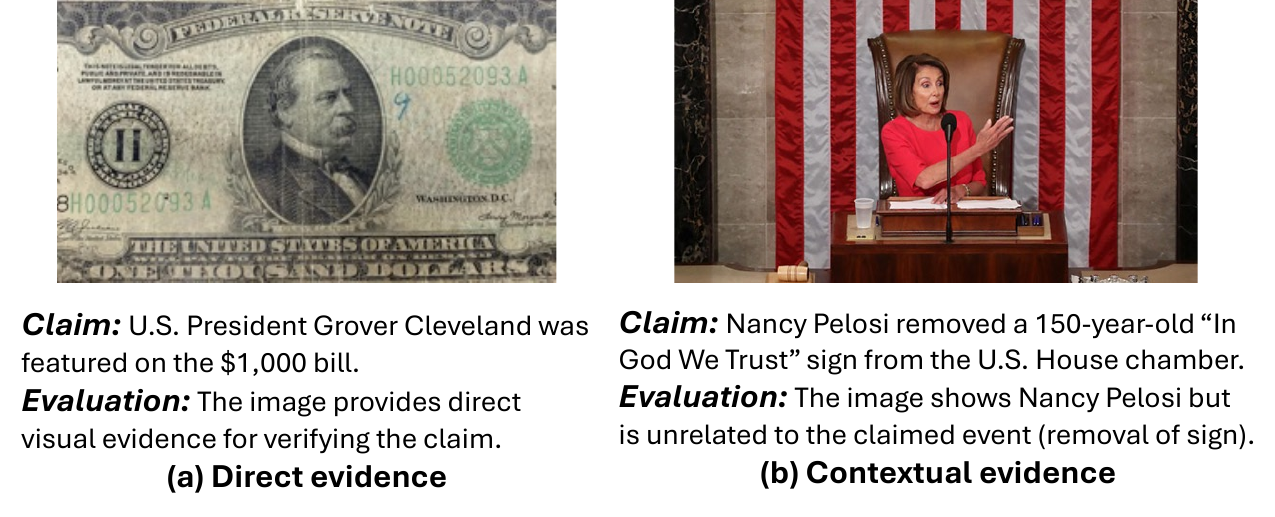}
    \caption{Examples of image annotations illustrating their evidential roles in claim verification, including \emph{direct evidence} and \emph{contextual evidence}.}
    \label{fig:image-categories}
\end{figure}

For images, annotators categorize their role based on their perceived evidential function within the multimodal claim context, rather than their factual correctness.
\emph{direct evidence}: the image is presented as providing direct and specific visual proof for the claim.
\emph{contextual evidence}: the image is semantically related to the claim and provides contextual information, but does not constitute direct visual proof.
\emph{irrelevant}: the image is unrelated to the claim.
Results show that images provide direct evidence in 87\% of claims, while only 8\% correspond to cases with irrelevant image usage.
Examples of these categories are shown in Fig.~\ref{fig:image-categories}.

\paragraph{Validation of Reasoning and Evidence}

We further validate the correctness of reasoning and evidence using the same rubric employed in our explainability evaluation.
Annotators assess whether (i) the reasoning logic and key points are supported by the linked evidence (reasoning hallucination), and (ii) the cited evidence is correctly used and accurately represented (evidence usage hallucination).
Detailed guidelines and aggregated results are reported in Sec.~\ref{exp-explain}.

%The inter-annotator agreement is 0.82 Cohen’s κ.
\subsection{Statistics} 
\begin{figure}[h]
    \centering
    \includegraphics[width=\columnwidth]{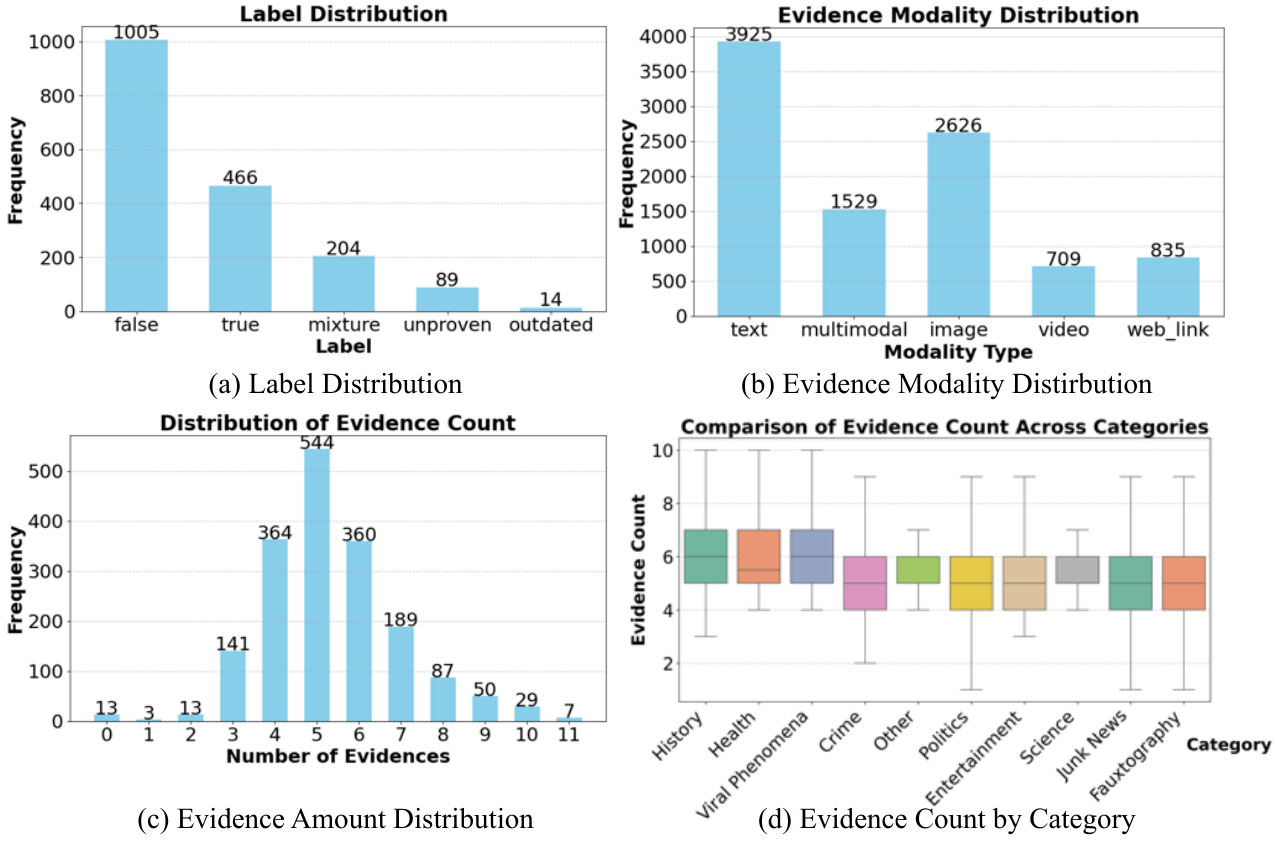}
    \caption{Statistics of RW-Post Dataset}
    \label{fig:rw_post_dataset_statistics}
\end{figure}

Summary statistics are shown in Fig.~\ref{fig:rw_post_dataset_statistics}.
The RW-Post dataset spans 10 domains, with Fauxtography (831), Politics (384), Entertainment (107), Junk News (99), and Viral Phenomena (64) being the most frequent categories. 

RW-Post’s main benchmark split contains 1.52k instances with three labels (True/False/Unproven). We additionally provide a challenge set of 218 Mixture/Outdated cases for future research on nuanced and time-sensitive verification (Fig.~\ref{fig:rw_post_dataset_statistics}a); we do not include this set in the main benchmark metrics unless otherwise stated.

Most articles contain diverse modalities of evidence such as text, image, and multimodal combinations (Fig.~\ref{fig:rw_post_dataset_statistics}b), and mostly contain 3–8 pieces of evidence (Fig.~\ref{fig:rw_post_dataset_statistics}c).
Categories like History, Health, and Viral Phenomena show higher median evidence counts (Fig.~\ref{fig:rw_post_dataset_statistics}d), indicating that human fact-checkers might conduct more in-depth investigations for these topics.

\section{Reference Verification Pipeline}
\label{method}

\paragraph{Overview}

To facilitate reproducible benchmarking on RW-Post, we provide \textbf{AgentFact}, a reference multimodal verification pipeline.
AgentFact decomposes open-web fact-checking into modular components for
(i) strategy planning,
(ii) textual evidence retrieval,
(iii) visual analysis via reverse image search, and
(iv) evidence-grounded reasoning and (v) explanation generation.

The pipeline supports both \textbf{open-web} evaluation (end-to-end evidence acquisition), enabling systematic auditing of evidence usage and explanation faithfulness.

\subsection{Problem Definition}

A multimodal claim is represented as $\{C, P\}$, where $C$ is the textual claim and $P=\{P_T,P_I\}$ denotes the post context consisting of post text and an associated image.

The goal is to determine the claim veracity and provide evidence-grounded reasoning.
The system outputs:

\begin{itemize}
\item a veracity label $V$,
\item a set of evidence items $E=\{(id_i, x_i, u_i)\}$ containing evidence snippets and source URLs,
\item a set of key reasoning points aligned with evidence $K_A$,
\item a reasoning summary explaining the verification process $R_A$,
\item and a confidence score $C_L$.
\end{itemize}

\textbf{Evidence Representation.}
Evidence is represented as items with stable identifiers,
$E=\{(id_i,x_i,u_i)\}$, where $x_i$ is a short evidence description and $u_i$ is the source URL.
Reasoning points cite evidence using these identifiers, enabling explicit evidence grounding and auditing.

\subsection{Agents for Fact-Checking}
\begin{figure*}
    \centering
\includegraphics[width=\linewidth]{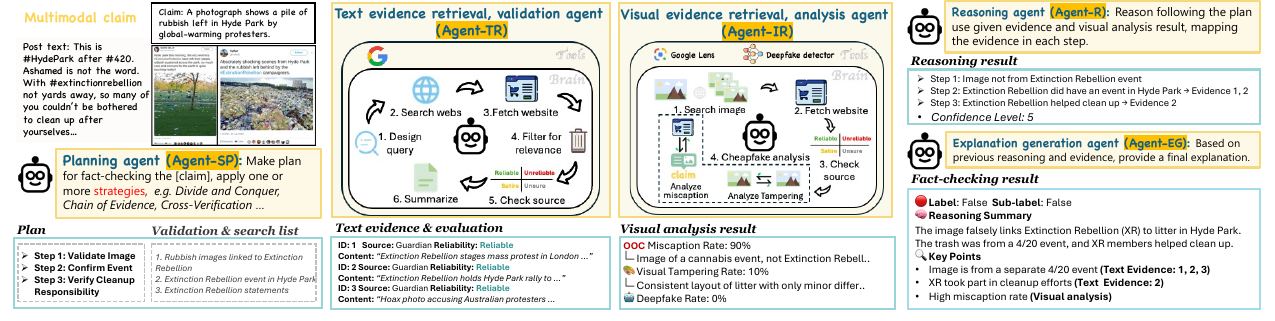}
    \caption{Reference pipeline components for open-web verification; used as baselines in our benchmark. Five agents are designed to handle five distinct subtasks, allowing task decomposition that reduces reasoning complexity. The iterative workflow is presented in Fig.~5.}
    \label{fig:agent}
\end{figure*}
As illustrated in Fig.~\ref{fig:agent}, the framework consists of five specialized agents:  
1) \textbf{Strategy Planning Agent (Agent-SP)} ,  
2) \textbf{Text Evidence Retrieval and Validation Agent (Agent-TR)},  
3) \textbf{Image Retrieval and Analysis Agent (Agent-IR)},  
4) \textbf{Reasoning Agent (Agent-R)},  
5) \textbf{Explanation Generation Agent (Agent-EG)}.
Each agent is designed with task-specific knowledge to mimic distinct components of real-world fact-checking workflows, and is deployed using prompt-engineering techniques with tool interaction. The simplified prompts are provided in the Appendix.

\subsubsection{Strategy Planning (Agent-SP)}
Agent-SP  takes as input the claim ($C$), the associated post ($P$), and—when applicable—previous reasoning trace. It generates a verification plan consisting of
(i) reasoning steps $S$,
(ii) a validation list $L_v$ containing statements to verify,
and (iii) a search list $L_s$ describing auxiliary search intents.

\subsubsection{Text Evidence Retrieval and Validation Agent (Agent-TR)}
Agent-TR retrieves textual evidence from the web to support claim verification.
The module operates through four steps:

\begin{enumerate}
    \item \textbf{Query Generation.}
    Search queries $Q$ are generated based on the search list $L_s$ and validation list $L_v$ produced by Agent-SP.

    \item \textbf{Web Search.}
    The Serper API\footnote{\url{https://serper.dev/}} is used to retrieve candidate webpages for each query.

    \item \textbf{Query-Guided Filtering.}
    Retrieved results are filtered and summarized to retain only evidence relevant to the input queries.

    \item \textbf{Source Reliability Estimation.}
    To reduce the influence of unreliable information, each source is classified into one of four categories: \textit{reliable}, \textit{unreliable}, \textit{satire}, or \textit{unsure}.
    The resulting reliability-annotated evidence is denoted as $E_r$.
\end{enumerate}
This design reduces the impact of misleading or satirical sources during reasoning.

\subsubsection{Image Retrieval and Analysis Agent (Agent-IR)}

Agent-IR analyzes the input image to detect potential visual inconsistencies in multimodal claims.
The module first performs reverse image search using the Google Cloud Vision API to retrieve visually similar images and their associated webpage text.

Retrieved images are categorized by an LLM into three types:
(1) \emph{near-duplicate images}, which are visually almost identical to the input image;
(2) \emph{same-event images}, which depict the same real-world event but differ in viewpoint or timing;
(3) \emph{unrelated images}, which are visually or semantically unrelated.

Images in the first two categories are retained as potentially informative evidence for verification.
Based on these results, Agent-IR performs three types of verification:
(i) \textbf{visual tampering detection}, by comparing the post image with retrieved images;
(ii) \textbf{miscaption detection}, by checking consistency between the claim/post text and retrieved webpage text;
(iii) \textbf{deepfake detection}, using the DIRE model~\cite{wang2023dire}.
The resulting signals are provided to the reasoning agent as auxiliary evidence.

\subsubsection{Reasoning Agent (Agent-R)}
The reasoning agent is provided with the claim ($C$), the post text ($P_T$), and the validation plan ($S^{(t)}$). 
It follows the prescribed reasoning steps, evaluating the relevant evidence at each step. Before the reasoning process begins, Agent-R is prompted to rephrase the input claim to improve focus and prevent the model from drifting away from the core verification target. 
The output ($R$) of Agent-R consists of three components:  
\textbf{Reasoning Results} — step-wise analysis, supporting evidence descriptions, and utility assessments;  
\textbf{Veracity Label} — the predicted truthfulness of the claim (e.g., true, false, unproven);  
\textbf{Confidence Level} — a score from 1 to 5 reflecting the model’s certainty.

\subsubsection{Explanation Generation Agent (Agent-EG)}
% Once the reasoning agent yields a high-confidence judgment, or after reaching a maximum reasoning depth, the explanation generation agent is triggered. 
The Explanation Generation Agent takes as input the reasoning outputs, together with the supporting textual and visual evidence, and produces a structured explanation. 
The output format follows the schema defined in the problem formulation.

\label{method}
\begin{figure}
    \centering
\includegraphics[width=\linewidth]{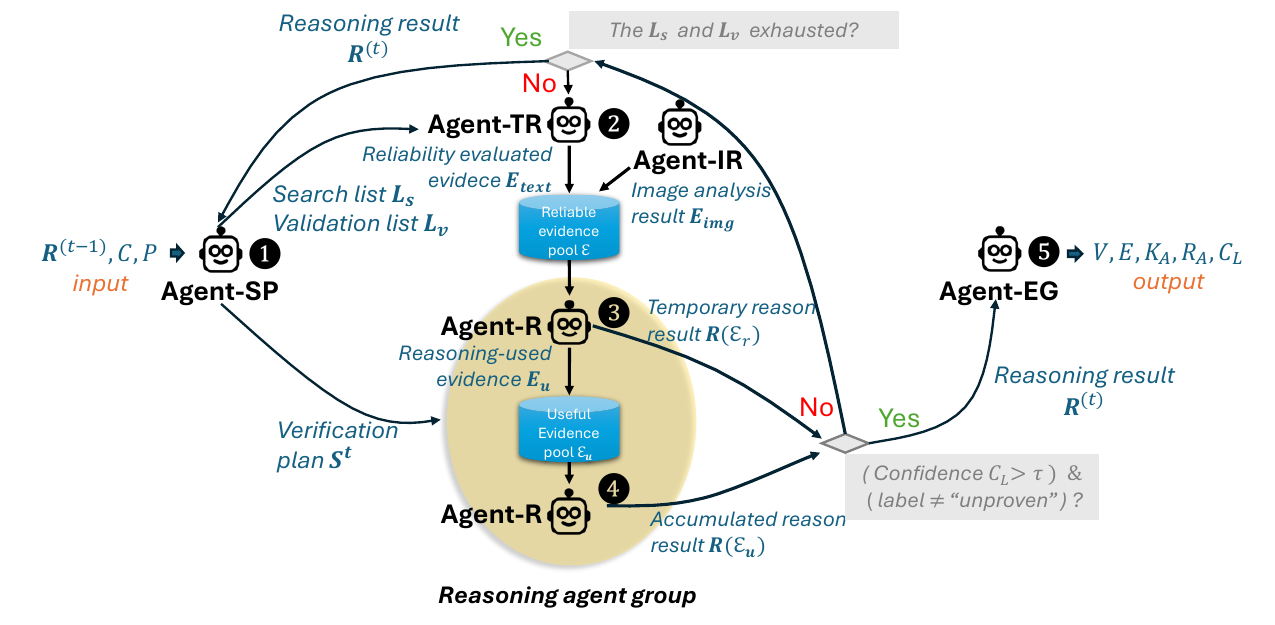}
\caption{Iterative retrieve–reason workflow of AgentFact.
The pipeline alternates between evidence acquisition and evidence-grounded reasoning, producing veracity predictions and explanations with explicit evidence-ID citations.}
    \label{fig:workflow}
\end{figure}

\subsection{Workflow}
\label{sec:workflow}
AgentFact is a reference verification framework which can operate in open-web verification mode.
As illustrated in Fig.~\ref{fig:workflow}, open-web verification follows an iterative retrieve–reason loop:

\begin{itemize}

\item \textbf{Step 1: Plan Generation.}
Agent-SP generates a verification plan $S^{(t)}$, together with a validation list $L_v$ (statements to verify) and a search list $L_s$ (auxiliary search intents).

\item \textbf{Step 2: Evidence Acquisition.}
Agent-TR retrieves textual evidence via web search, while Agent-IR performs reverse image search and visual analysis.
The retrieved results are combined into an evidence pool $\mathcal{E}$.

\item \textbf{Step 3: Evidence Selection.}
Agent-R performs intermediate reasoning to identify evidence relevant to the current verification plan, forming a useful evidence subset $\mathcal{E}_u$.

\item \textbf{Step 4: Evidence-Grounded Verification.}
Agent-R performs reasoning over $\{C,P,\mathcal{E}_u\}$ to produce a veracity prediction and a confidence score.
If confidence is low or the claim remains unverified, another retrieval–reason iteration is triggered.

\item \textbf{Step 5: Explanation Generation.}
Agent-EG organizes the reasoning results and supporting evidence into a structured explanation with explicit evidence-ID citations.

\end{itemize}

\section{Experiments}
\label{sec:experiment}

\paragraph{Evaluation dimensions}
Performance is measured on (i) \textbf{veracity prediction} and (ii) \textbf{evidence-grounded explainability}, which assesses whether reasoning and key points are supported by verifiable evidence.

\paragraph{Datasets}
RW-Post is our primary benchmark for evaluating evidence-grounded multimodal fact-checking.
We additionally evaluate on Mocheg~\cite{mocheg23}, CLAIMREVIEW2024+ (CR2024+)~\cite{braun2024defame}, and the OOC dataset NewsCLIPpings to test generalization~\cite{luo2021newsclippings}.
Due to the cost of search-based methods, we sample 1,000 test instances for Mocheg and NewsCLIPpings, while using the full test sets for RW-Post and CR2024+.
We follow each dataset’s official label space (binary for OOC and three-class for fact-checking).

\paragraph{Baselines}
We compare AgentFact with three categories of baselines:
(1) feature-based OOC detectors (CCN~\cite{abdelnabi2022open}, Sniffer~\cite{qi2024sniffer}),
(2) retrieval-augmented LLM/LVLM fact-checking systems (LEMMA~\cite{xuan2024lemma}, DEFAME~\cite{braun2024defame}),
and (3) open-source LVLMs (LLaVA-1.5 and Qwen2-VL-Chat, \emph{7B-scale}) and strong close-source LVLM (GPT-4o-mini).

\paragraph{Evaluation Regimes}
We evaluate models under the three benchmark regimes defined in Sec.~\ref{sec:dataset}:
\textbf{closed-book}, \textbf{evidence-bounded}, and \textbf{open-web}.
Search-based systems (AgentFact, LEMMA, and DEFAME) are evaluated in the open-web setting,
while LVLMs are evaluated in the closed-book and evidence-bounded settings on RW-Post.

\subsection{Implementation Details}
\label{sec:implementation_details}
\textbf{Model and API Settings}.
For search-based systems (LEMMA, DEFAME, and AgentFact), we conduct an inference-only evaluation and standardize key components. We use GPT-4o-mini as the underlying LLM and the Serper API as the unified web search backend with a fixed top-$10$ results per query.
% LVLM baselines (LLaVA-1.5 and Qwen2-VL-Chat) are evaluated using their native models without additional LLM calls.
For all open-source LVLM baselines (LLaVA-1.5 and Qwen2-VL-Chat), we use their official checkpoints and default image processors.
Specifically, we evaluate them with greedy decoding (\texttt{do\_sample=False}) and set the maximum generation length to 256 tokens.
% Each test instance is processed independently without conversation history.
For Sniffer, we use the publicly released pretrained checkpoints trained on NewsCLIPpings.
For CCN, we report the results from the original paper, as reproducing the training requires access to the full dataset and preprocessing pipeline.

\textbf{Computing Resources}.
Search-based systems are run on a CPU-only machine.
LVLM baselines are evaluated on a single NVIDIA H100 GPU.

\textbf{Evidence Settings}.
In evidence-bounded evaluation, models are provided with the linked RW-Post evidence items.
In open-web evaluation, all retrieved URLs and evidence snippets are logged for auditing.

\textbf{Data Leakage Prevention}.
To prevent leakage, we filter evidence by excluding
(i) results from major fact-checking websites (e.g., Snopes) and
(ii) sources published on or after the original fact-checking date.

\subsection{Open-web End-to-End Veracity Results} 
\paragraph{Metrics}
We evaluate veracity prediction under two settings: 
\textbf{binary classification} for OOC detection and 
\textbf{three-class classification} (\texttt{True}, \texttt{False}, \texttt{Unproven}) for fact-checking. 
For both settings we report \textbf{Macro-averaged F1 (Macro-F1)}, \textbf{Weighted F1}, and \textbf{Accuracy (Acc)}. 
Macro-F1 is our primary metric to account for label imbalance. 
All metrics are reported as percentages (multiplied by 100).

\begin{table}[t]
\caption{
Veracity classification results across diverse datasets. 
``OOC'': 2-class out-of-context.
``FC'': 3-class fact-checking.
}
\label{tab:veracity_results}
\centering
\small
\setlength{\tabcolsep}{4pt}
\begin{tabular}{p{1.5cm}ccccc}
\toprule
Dataset & Task & Model & Macro-F1 & Weighted-F1 & Acc \\
\midrule
\multirow{5}{*}{\makecell{NewsCLIP-\\pings}} & \multirow{5}{*}{OOC}
  & CCN & -- & -- & 84.7 \\
  & & Sniffer & \textbf{86.4} & \textbf{86.5} & \textbf{86.5} \\
  & & LEMMA & 51.1 & 52.3 & 58.2 \\
  & & DEFAME & 60.3 & 60.7 & 60.8 \\
  & & \textbf{AgentFact} & 56.0 & 56.6 & 59.7 \\
\midrule
\multirow{3}{*}{Mocheg} & \multirow{3}{*}{FC}
  & LEMMA & 26.2 & 28.0 & 37.5 \\
  & & DEFAME & 44.4 & 44.9 & 46.0 \\
  & & \textbf{AgentFact}& \textbf{49.7} & \textbf{51.9} & \textbf{48.2} \\
\midrule
\multirow{3}{*}{CR2024+} & \multirow{3}{*}{FC}
  & LEMMA & 15.0 & 16.5 & 14.5 \\
  & & DEFAME & 29.6 & 59.3 & 52.1 \\
  & & \textbf{AgentFact}& \textbf{50.7} & \textbf{69.2} & \textbf{67.0} \\
\midrule
\multirow{3}{*}{RW-Post} & \multirow{3}{*}{FC}
  & LEMMA & 39.9 & 45.6 & 42.1 \\
  & & DEFAME & 43.4 & 51.4 & 46.5 \\
  & & \textbf{AgentFact}& \textbf{48.7} & \textbf{65.5} & \textbf{57.8} \\
\bottomrule
\end{tabular}
\end{table}

\paragraph{Main results}
Table~\ref{tab:veracity_results} reports end-to-end \textbf{open-web} veracity results under a unified search backend with leakage-controlled filtering.
AgentFact achieves the best performance among fact-checking systems on the three fact-checking datasets in terms of Macro-F1, Weighted-F1 and Acc, while specialized OOC detectors (CCN and Sniffer) remain strongest on the OOC task. 

\paragraph{Fact-Checking (3-class)}
AgentFact consistently improves both Macro-F1 and Weighted-F1 over LEMMA and DEFAME on RW-Post and ClaimReview2024+, indicating that stronger evidence acquisition and evidence-grounded reasoning benefit complex real-world verification.
On Mocheg, which lacks post-image context, AgentFact still achieves clear gains, suggesting that the pipeline generalizes to text-only verification.
Notably, improvements in Macro-F1 demonstrate that AgentFact better handles class imbalance, particularly for minority classes.

\paragraph{OOC Detection (binary)}
On NewsCLIPpings, task-specific detectors outperform general-purpose fact-checking systems across all metrics, as expected.
This gap highlights the difference between OOC detection and open-web fact-checking, motivating the controlled RW-Post benchmark (Sec.~\ref{sec:controlled_benchmark}) for deeper analysis of evidence utilization and visual grounding.

\subsection{Controlled Benchmarking on RW-Post (LVLM Baselines)}
\label{sec:controlled_benchmark}
To enable controlled and reproducible evaluation, we evaluate LVLMs on the RW-Post dataset under two regimes: 
(1) \textbf{closed-book} and (2) \textbf{evidence-bounded}. 
Specifically, we consider four input configurations--\textit{T}, \textit{T+I}, \textit{T+E}, and \textit{T+I+E}--to examine the impact of incorporating image (I) and external evidence (E) in the model input.

\begin{table}[t]
\centering
\small
\setlength{\tabcolsep}{3pt}
\caption{
Controlled RW-Post evaluation.
$\mathbf{T}$: text; $\mathbf{I}$: image; $\mathbf{E}$: evidence.
All models are evaluated on the same set of instances (N=1268).
Format denotes the percentage of outputs that follow the required structured format.
}
\label{tab:lvlm_rwpost}
\resizebox{\linewidth}{!}{
\begin{tabular}{ccccccc}
\toprule
Category & Model & Setting  & Macro-F1 & Weighted-F1 & Acc & FSR (\%) \\
\midrule
\multirow{8}{*}{\makecell{Open-\\source\\LLM}} 
& \multirow{4}{*}{Qwen2-VL-Chat} 
& T      & 22.0 & 18.2 & 30.9 & 50.3 \\
& & T+I    & 18.9 & 18.7 & 30.3 & 50.3 \\
& & T+E    & 35.2 & 42.3 & 44.0 & 50.3 \\
& & T+I+E & 32.3 & 35.5 & 40.8 & 50.3 \\
\cmidrule(lr){2-7}
& \multirow{4}{*}{LLaVA-1.5} 
& T      & 21.1 & 16.8 & 24.7 & 86.2 \\
& & T+I    & 22.4 & 18.0 & 26.9 & 86.2 \\
& & T+E    & 40.1 & 47.9 & 47.9 & 86.2 \\
& & T+I+E  & 38.2 & 45.0 & 47.3 & 86.2 \\
\midrule
\multirow{4}{*}{\makecell{Close-\\source\\LLM}}
& \multirow{4}{*}{GPT-4o-mini} 
& T      & 26.3 & 25.4 & 30.6 & 100.0 \\
& & T+I    & 30.4 & 30.7 & 34.5 & 100.0 \\
& & T+E  & 67.8 & \textbf{85.1} & \textbf{84.5} & 100.0 \\
& & T+I+E  & \textbf{68.8} & 84.5 & 83.8  & 100.0 \\
% \midrule
% Agent
% & \textbf{AgentFact} & T+E  & 65.8 & \textbf{86.3} & \textbf{85.5} & 99.9 \\
\bottomrule
\end{tabular}
}
\end{table}

\begin{table}[t]
\centering
\small
\setlength{\tabcolsep}{2pt}
\caption{
Per-class recall and F1 on RW-Post under controlled evaluation. 
We report (i) closed-book T and (ii) evidence-bounded T+E. 
All values are percentages.
}
\label{tab:perclass_recall_f1}
\begin{tabular}{p{2.1cm}p{0.9cm}cccccc}
\toprule
& & \multicolumn{3}{c}{Recall} & \multicolumn{3}{c}{F1} \\
\cmidrule(lr){3-5} \cmidrule(lr){6-8}
Model & Setting 
& R$_\text{False}$ & R$_\text{True}$ & R$_\text{Unprov}$ 
& F1$_\text{False}$ & F1$_\text{True}$ & F1$_\text{Unprov}$ \\
\midrule

\multirow{2}{*}{Qwen2-VL-Chat}
& T   
& 3.8  & 95.5 & 9.4  
& 7.2 & 48.6 & 7.6 \\
& T+E 
& 24.1 & 91.6 & 20.5 
& 38.7 & 53.2 & 12.2 \\

\midrule

\multirow{2}{*}{LLaVA-1.5} 
& T   
& 2.2  & 71.4 & 46.0 
& 4.4 & 45.9 & 12.3 \\
& T+E 
& 27.7 & \textbf{97.8} & 28.6 
& 43.0 & 66.5 & 10.6 \\

\midrule

\multirow{2}{*}{GPT-4o-mini}
& T   
& 7.5  & 76.2 & \textbf{49.3} 
& 13.9 & 52.0 & 13.1 \\
& T+E 
& 84.0 & 95.6 & 31.0 
& 89.4 & 86.7 & \textbf{27.3} \\

\bottomrule
\end{tabular}
\end{table}

\paragraph*{Main results}

All models are evaluated on the same set of RW-Post instances (N=1268). 
For outputs that fail to follow the required structured format, we additionally report the Format Success Rate (FSR), which measures each model’s ability to produce valid structured outputs (Table~\ref{tab:lvlm_rwpost}). 
The other metrics are computed only on outputs that conform to the required format.

\textbf{Evidence substantially improves performance.}
\textit{Open-source LVLMs:} Both LLaVA-1.5 and Qwen2-VL-Chat show large gains when evidence is provided (T$\rightarrow$T+E). 
For example, Qwen2-VL-Chat improves from 18.2 to 42.3 (Weighted-F1), and LLaVA-1.5 improves from 16.8 to 47.9. 
Similar trends are observed in Macro-F1, indicating improved performance across classes rather than only majority-class gains.

\textit{Closed-source LLM:} GPT-4o-mini benefits even more from evidence, improving from 25.4 to 85.1 (Weighted-F1), with corresponding gains in Macro-F1 (26.3$\rightarrow$67.8). 
This confirms that evidence-grounded reasoning is the dominant factor for RW-Post performance, and that the benchmark is not reliably solvable in a closed-book setting.

\textbf{Visual grounding remains challenging.}
Adding images yields limited or negative gains. 
In closed-book settings (T$\rightarrow$T+I), improvements are marginal across models. 
In evidence-bounded settings (T+E$\rightarrow$T+I+E), performance often decreases, e.g., Qwen2-VL-Chat (42.3$\rightarrow$35.5 Weighted-F1), LLaVA-1.5 (47.9$\rightarrow$45.0), and GPT-4o-mini (85.1$\rightarrow$84.5), suggesting that current LVLMs struggle to effectively integrate visual signals with textual evidence.

\textbf{Refutation and uncertainty remain difficult.}
Per-class results (Table~\ref{tab:perclass_recall_f1}) show that FALSE recall remains relatively low compared to TRUE, indicating that models are better at confirming supported claims than rejecting misinformation. 
Performance on Unproven cases remains unstable, reflected by consistently lower F1 scores, highlighting challenges in calibrated abstention under inconclusive evidence.

\textbf{Format compliance varies significantly across models.}
Open-source LVLMs show substantially lower format success rates (50.3\% for Qwen2-VL-Chat and 86.2\% for LLaVA-1.5) compared to GPT-4o-mini (100\%). 
This indicates that structured output generation is itself a non-trivial challenge, and directly impacts the practical usability of models in evidence-grounded fact-checking pipelines.

\textbf{Open-source LVLMs lag behind strong closed-source baselines in evidence utilization.}
Even under evidence-bounded evaluation, open-source LVLMs remain far below GPT-4o-mini (roughly 35--48 Weighted-F1 vs.\ 84--85), suggesting a significant gap in evidence comprehension, contradiction handling, and decision-making grounded in external evidence.

\subsection{Evidence Grounding and Explainability}
\label{exp-explain}

We evaluate explanation quality through human assessment following prior LLM-based fact-checking studies~\cite{kim2024can, zhang2023towards}.
Forty RW-Post instances are randomly sampled by 3 annotators.
Outputs from models and RW-Post labels are presented in randomized order with anonymized identifiers.

\paragraph{Annotation Scheme}
Annotators evaluate each output along three dimensions:
(1) \textbf{Reasoning Hallucination} (unsupported or logically invalid inference), 
(2) \textbf{Evidence Usage Hallucination} (misused, irrelevant, or fabricated evidence), and 
(3) \textbf{Label Justification} (whether the predicted label is well-supported).
Each dimension is scored on a three-level ordinal scale (0–2), where lower scores indicate fewer errors and higher faithfulness. 
For presentation, scores are normalized to the range $[0,1]$.

Detailed annotation guidelines and the interface are provided in Appendix.

\paragraph{Results}
Table~\ref{tab:human-eval} reports the average scores across three annotators.
(1) Ground-truth rationales achieve near-perfect scores (0.94–1.00), demonstrating high faithfulness and serving as a strong upper bound for explanation quality.
(2) AgentFact significantly outperforms LEMMA across all criteria.
Using the Wilcoxon signed-rank test on paired differences, we obtain $p<10^{-5}$ for reasoning, evidence usage, and label justification.
Bootstrap confidence intervals (95\%) further confirm the robustness of these improvements.

\begin{table}[thb]
\setlength{\tabcolsep}{5pt}
\centering
% \caption{Average human evaluation scores (0–1). 
% Right columns report paired significance between AgentFact and LEMMA using original 0–2 ratings. 
% $\Delta$: mean paired difference (AgentFact $-$ LEMMA).}
\caption{Average human evaluation scores (0–1).
Right columns report paired significance between AgentFact and LEMMA.
$\Delta$: mean paired difference (AgentFact $-$ LEMMA).}
\label{tab:human-eval}
\begin{tabular}{cccc|cc}
\toprule
\textbf{Criteria} & \textit{\textbf{GT}} & \textbf{LEMMA} & \textbf{AgentFact} 
& $\Delta$ & \textbf{p-value} \\
\midrule
Reasoning H. \scriptsize{↑}  & 0.969    
& 0.445 & \textbf{0.767}
& 0.32 & $6.8\times10^{-6}$ \\

Evidence H. \scriptsize{↑} & 1.000   
& 0.796 & \textbf{0.946}  
& 0.15 & $9.1\times10^{-6}$ \\

Label J. \scriptsize{↑} & 0.937     
& 0.429 & \textbf{0.758} 
& 0.33 & $1.1\times10^{-6}$ \\
\bottomrule
\end{tabular}
\end{table}

\subsection{Ablation Study of Agents}
Table~\ref{tab:ablation-results} reports the ablation results across the four agent groups.\footnote{The ablation study is conducted on an earlier version of AgentFact without the evidence quality assessment. Therefore, the absolute performance is not directly comparable to the main results. We instead focus on relative performance differences to analyze the contribution of each agent.}

\paragraph{Overall Trends.}
The full AgentFact system achieves the best performance across all metrics, confirming the complementary contributions of different agent modules. 
Among them, text retrieval (Agent-TR) is the most critical component: removing it leads to the largest performance drop (weighted F1: 0.629 $\rightarrow$ 0.469; accuracy: 0.570 $\rightarrow$ 0.373).

\paragraph{Effect of Individual Agents.}
Removing the strategy planning agent (Agent-SP) results in moderate degradation (weighted F1: $-3.6\%$; accuracy: $-4.3\%$), indicating that structured planning improves evidence retrieval efficiency. 
Similarly, removing the temporary reasoning agent (Agent-R-I, Agent-R that performs intermediate reasoning) slightly reduces performance (weighted F1: $-1.8\%$; recall: $-2.9\%$), suggesting its role in filtering noisy evidence.
The removal of the image retrieval and analysis agent (Agent-IR) leads to comparable declines (weighted F1: $-2.6\%$; recall: $-3.6\%$), highlighting the contribution of visual signals in multimodal verification.

\begin{table}[h]
\centering
\caption{Ablation study results on different agents}
\begin{tabular}{lcccccc}
\toprule
Model & Macro-F1 & Weighted-F1 & Acc & P  & Rec  \\
\midrule
\textit{w/o Agent SP }      & 0.470  &  0.593& 0.527 & 0.766  & 0.527                      \\
\textit{w/o Agent-IR}       & 0.468  &  0.603    &       0.534            &     0.775           &     0.534  \\
\textit{w/o Agent-TR}       & 0.328  &  0.469 & 0.373 &   0.763 &  0.373              \\
\textit{w/o Agent-R-I}      & 0.473  & 0.611 & 0.541 & \textbf{0.775} & 0.541 \\
\midrule
\textit{Full System}        & \textbf{0.489} &  \textbf{0.629}  & \textbf{0.570} & 0.770 & \textbf{0.570}\\
\bottomrule
\end{tabular}
\label{tab:ablation-results}
\end{table}

\subsection{Case Study}

\begin{figure}[H]
    \centering

    % ====================
    % Top: Image | Claim (Left–Right)
    % ====================
    \begin{minipage}[c]{0.36\linewidth}
        \centering
        \includegraphics[width=\linewidth]{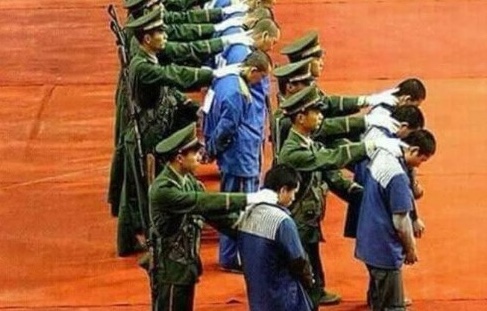}
    \end{minipage}
    \hfill
    \begin{minipage}[c]{0.60\linewidth}
        \footnotesize
        \textbf{\textsc{Claim:}}\\
        A photograph shows 26 corrupt politicians shortly before they were executed
        in China.
    \end{minipage}

    \vspace{0.8em}
\setlength{\tabcolsep}{3pt}
\renewcommand{\arraystretch}{1.05}
\footnotesize
\begin{tabularx}{\linewidth}{X}
\toprule

\textsc{Label} \quad 
\textbf{Model:} \textsc{False} \quad 
\textbf{Human:} \textsc{False} \\

\midrule
\textsc{Reasoning} \\
\textbf{Model:}
The image is entirely miscaptioned and unrelated to the alleged execution; no evidence supports the claim.\\
\textbf{Human:}
The image shows a 2004 sentencing rally in Wenzhou, depicting convicted criminals rather than politicians. \\

\midrule
\textsc{Key Points} \\
\textbf{Model:}
(1) 100\% miscaptioned; (2) no supporting evidence.\\
\textbf{Human:}
(1) 2004 Wenzhou sentencing rally; (2) subjects are convicted criminals. \\

\midrule
\textsc{Evidence} \\
\textbf{Model:}
[1] Getty Images search result titled ``3,619 Death Row Inmate Stock Photos'' $\Rightarrow$ \hlred{not depicting politicians (100\% miscaption)}; 
[2] Human Rights Watch: general background on China’s anti-corruption campaign, lacking any direct reference to the image or event.\\
\textbf{Human:}
[1] Getty Images: original caption confirming sentencing rally of \hlgreen{convicted criminals} (2004, Wenzhou). \\

\midrule
\textsc{Comparison} \\
\textbf{Model relies on coarse miscaption signals, while human reasoning identifies the exact event and context.} \\

\bottomrule
\end{tabularx}

\caption{
Case study of a correctly classified claim with weak evidence and reasoning.
}
\label{fig:china_execution_vertical_case}
\end{figure}

We present a representative case study in Fig.~\ref{fig:china_execution_vertical_case}. 
This example shows that AgentFact can generate coherent reasoning and structured key points grounded in multimodal evidence. 
This case highlights a common failure pattern, which we term \textbf{weak contextual grounding}, where retrieved evidence appears relevant but fails to capture the true event-level context of the image.
In practice, LVLM's miscaption detection tends to rely on surface-level textual matching between retrieved content and the claim, without assessing semantic relevance. 
As a result, non-informative content (e.g., advertisements) may be incorrectly treated as valid evidence.
Similar patterns are consistently observed across other cases (see Appendix).

\section{Conclusion}

We introduce \textbf{RW-Post}, a benchmark for evidence-grounded multimodal fact-checking that aligns real-world claims with their original posts, structured reasoning traces, and explicitly linked evidence. 
RW-Post enables controlled evaluation of multimodal verification under complementary regimes. 
Experiments with strong LVLM baselines show that reliable evidence substantially improves verification performance, while visual grounding and robust misinformation refutation remain challenging. 
We hope RW-Post provides a useful testbed for advancing research on evidence-aware multimodal fact-checking. 
%%
%% The acknowledgments section is defined using the "acks" environment
%% (and NOT an unnumbered section). This ensures the proper
%% identification of the section in the article metadata, and the
%% consistent spelling of the heading.
\section*{Acknowledgments}
This research / project is supported by the Ministry of Education, Singapore, under its MOE AcRF TIER 3 Grant (MOE-MOET32022-0001).
Any opinions, findings and conclusions or recommendations expressed in this material are those of the author(s) and do not reflect the views of MOE, Singapore.

%%
%% The next two lines define the bibliography style to be used, and
%% the bibliography file.
\bibliographystyle{IEEEtran}
\bibliography{samples/main}

@String{Computing = "Computing" }

@String{Computer = "{IEEE} Computer" }

@String{Springer = "Springer-Verlag" }

@article{kim2024can,
  title={Can LLMs Produce Faithful Explanations For Fact-checking? Towards Faithful Explainable Fact-Checking via Multi-Agent Debate},
  author={Kim, Kyungha and Lee, Sangyun and Huang, Kung-Hsiang and Chan, Hou Pong and Li, Manling and Ji, Heng},
  journal={arXiv preprint arXiv:2402.07401},
  year={2024}
}

@inproceedings{braun2024defame,
   title = {{DEFAME: Dynamic Evidence-based FAct-checking with Multimodal Experts}}, 
   author = {Tobias Braun and Mark Rothermel and Marcus Rohrbach and Anna Rohrbach},
   booktitle = {Proceedings of the 42nd International Conference on Machine Learning},
   year = {2025},
   url = {https://arxiv.org/abs/2412.10510},
}

@article{zhang2023towards,
  title={Towards llm-based fact verification on news claims with a hierarchical step-by-step prompting method},
  author={Zhang, Xuan and Gao, Wei},
  journal={AACL},
  year={2023}
}

@article{chen2024metasumperceiver,
  title={MetaSumPerceiver: Multimodal Multi-Document Evidence Summarization for Fact-Checking},
  author={Chen, Ting-Chih and Tang, Chia-Wei and Thomas, Chris},
  journal={ACL},
  year={2024}
}

@inproceedings{kareem2023fighting,
  title={Fighting Lies with Intelligence: Using Large Language Models and Chain of Thoughts Technique to Combat Fake News},
  author={Kareem, Waleed and Abbas, Noorhan},
  booktitle={International Conference on Innovative Techniques and Applications of Artificial Intelligence},
  pages={253--258},
  year={2023},
  organization={Springer}
}

@article{leite2023detecting,
  title={Detecting misinformation with llm-predicted credibility signals and weak supervision},
  author={Leite, Jo{\~a}o A and Razuvayevskaya, Olesya and Bontcheva, Kalina and Scarton, Carolina},
  journal={arXiv preprint arXiv:2309.07601},
  year={2023}
}

@inproceedings{flamingo-2022,
author = {Alayrac, Jean-Baptiste and Donahue, Jeff and Luc, Pauline and Miech, Antoine and Barr, Iain and Hasson, Yana and Lenc, Karel and Mensch, Arthur and Millicah, Katie and Reynolds, Malcolm and Ring, Roman and Rutherford, Eliza and Cabi, Serkan and Han, Tengda and Gong, Zhitao and Samangooei, Sina and Monteiro, Marianne and Menick, Jacob and Borgeaud, Sebastian and Brock, Andrew and Nematzadeh, Aida and Sharifzadeh, Sahand and Binkowski, Mikolaj and Barreira, Ricardo and Vinyals, Oriol and Zisserman, Andrew and Simonyan, Karen},
title = {Flamingo: a visual language model for few-shot learning},
year = {2022},
isbn = {9781713871088},
publisher = {Curran Associates Inc.},
address = {Red Hook, NY, USA},
booktitle = {Proceedings of the 36th International Conference on Neural Information Processing Systems},
articleno = {1723},
numpages = {21},
location = {New Orleans, LA, USA},
series = {NIPS '22}
}

@inproceedings{yang-etal-2022-coarse,
    title = "A Coarse-to-fine Cascaded Evidence-Distillation Neural Network for Explainable Fake News Detection",
    author = "Yang, Zhiwei  and
      Ma, Jing  and
      Chen, Hechang  and
      Lin, Hongzhan  and
      Luo, Ziyang  and
      Chang, Yi",
    editor = "Calzolari, Nicoletta  and
      Huang, Chu-Ren  and
      Kim, Hansaem  and
      Pustejovsky, James  and
      Wanner, Leo  and
      Choi, Key-Sun  and
      Ryu, Pum-Mo  and
      Chen, Hsin-Hsi  and
      Donatelli, Lucia  and
      Ji, Heng  and
      Kurohashi, Sadao  and
      Paggio, Patrizia  and
      Xue, Nianwen  and
      Kim, Seokhwan  and
      Hahm, Younggyun  and
      He, Zhong  and
      Lee, Tony Kyungil  and
      Santus, Enrico  and
      Bond, Francis  and
      Na, Seung-Hoon",
    booktitle = "Proceedings of the 29th International Conference on Computational Linguistics",
    month = oct,
    year = "2022",
    address = "Gyeongju, Republic of Korea",
    publisher = "International Committee on Computational Linguistics",
    url = "https://aclanthology.org/2022.coling-1.230",
    pages = "2608--2621",
}

@inproceedings{PanQACheck23,
  author       = {Liangming Pan and Xinyuan Lu and Min-Yen Kan and Preslav Nakov},
  title        = {QACHECK: A Demonstration System for Question-Guided Multi-Hop Fact-Checking},
  booktitle    = {Proceedings of the 2023 Conference on Empirical Methods in Natural Language Processing System Demonstrations Track (EMNLP 2023 Demo Track)},
  address      = {Singapore},
  year         = {2023},
  month        = {Dec}
}

@inproceedings{nan2021mdfend,
  title={MDFEND: Multi-domain Fake News Detection},
  author={Nan, Qiong and Cao, Juan and Zhu, Yongchun and Wang, Yanyan and Li, Jintao},
  booktitle={Proceedings of the 30th ACM International Conference on Information \& Knowledge Management},
  pages={3343--3347},
  year={2021}
}

@inproceedings{weibo_twitter,
author = {Jin, Zhiwei and Cao, Juan and Guo, Han and Zhang, Yongdong and Luo, Jiebo},
title = {Multimodal Fusion with Recurrent Neural Networks for Rumor Detection on Microblogs},
year = {2017},
isbn = {9781450349062},
publisher = {Association for Computing Machinery},
address = {New York, NY, USA},
url = {https://doi.org/10.1145/3123266.3123454},
doi = {10.1145/3123266.3123454},
booktitle = {Proceedings of the 25th ACM International Conference on Multimedia},
pages = {795–816},
numpages = {22},
location = {Mountain View, California, USA},
series = {MM '17}
}

@article{shu2020fakenewsnet,
  title={Fakenewsnet: A data repository with news content, social context, and spatiotemporal information for studying fake news on social media},
  author={Shu, Kai and Mahudeswaran, Deepak and Wang, Suhang and Lee, Dongwon and Liu, Huan},
  journal={Big data},
  volume={8},
  number={3},
  pages={171--188},
  year={2020},
  publisher={Mary Ann Liebert, Inc., publishers 140 Huguenot Street, 3rd Floor New~…}
}

@article{nakamura2020r,
	title={Fakeddit: A new multimodal benchmark dataset for fine-grained fake news detection},
	author={Nakamura, Kai and Levy, Sharon and Wang, William Yang},
	journal={Conference on Language Resources and Evaluation (LREC 2020)},
	pages={6149--6157},
	year={2020}
}

@inproceedings{przybyla2020capturing,
  title={Capturing the style of fake news},
  author={Przybyla, Piotr},
  booktitle={Proceedings of the AAAI conference on artificial intelligence},
  volume={34},
  number={01},
  pages={490--497},
  year={2020}
}

@inproceedings{zhou2020multimodal,
  title={SAFE: Similarity-Aware Multi-modal Fake News Detection},
  author={Zhou, Xinyi and Wu, Jindi and Zafarani, Reza},
  booktitle={Pacific-Asia Conference on Knowledge Discovery and Data Mining},
  pages={354--367},
  year={2020},
  organization={Springer}
}

@inproceedings{shu2020hierarchical,
  title={Hierarchical propagation networks for fake news detection: Investigation and exploitation},
  author={Shu, Kai and Mahudeswaran, Deepak and Wang, Suhang and Liu, Huan},
  booktitle={Proceedings of the international AAAI conference on web and social media},
  volume={14},
  pages={626--637},
  year={2020}
}

@inproceedings{NielsenMcConville2022,
  title = {MuMiN: A Large-Scale Multilingual Multimodal Fact-Checked Misinformation Social Network Dataset},
  author = {Dan Saattrup Nielsen and Ryan McConville},
  booktitle = {Proceedings of the 45th International ACM SIGIR Conference on Research and Development in Information Retrieval (SIGIR)},
  year = {2022},
  publisher = {ACM},
  eprint = {2202.11684}
}

@inproceedings{mocheg23,
author = {Yao, Barry Menglong and Shah, Aditya and Sun, Lichao and Cho, Jin-Hee and Huang, Lifu},
title = {End-to-End Multimodal Fact-Checking and Explanation Generation: A Challenging Dataset and Models},
year = {2023},
isbn = {9781450394086},
publisher = {Association for Computing Machinery},
address = {New York, NY, USA},
url = {https://doi.org/10.1145/3539618.3591879},
doi = {10.1145/3539618.3591879},
booktitle = {Proceedings of the 46th International ACM SIGIR Conference on Research and Development in Information Retrieval},
pages = {2733–2743},
numpages = {11},
location = {Taipei, Taiwan},
series = {SIGIR '23}
}

@inproceedings{mishra2022factify,
  title={FACTIFY: A Multi-Modal Fact Verification Dataset.},
  author={Mishra, Shreyash and Suryavardan, S and Bhaskar, Amrit and Chopra, Parul and Reganti, Aishwarya N and Patwa, Parth and Das, Amitava and Chakraborty, Tanmoy and Sheth, Amit P and Ekbal, Asif and others},
  booktitle={DE-FACTIFY@ AAAI},
  year={2022}
}

@inproceedings{hu2023mr2,
  title={Mr2: A benchmark for multimodal retrieval-augmented rumor detection in social media},
  author={Hu, Xuming and Guo, Zhijiang and Chen, Junzhe and Wen, Lijie and Yu, Philip S},
  booktitle={Proceedings of the 46th international ACM SIGIR conference on research and development in information retrieval},
  pages={2901--2912},
  year={2023}
}

@article{liu2024mmfakebench,
  title={MMFakeBench: A Mixed-Source Multimodal Misinformation Detection Benchmark for LVLMs},
  author={Liu, Xuannan and Li, Zekun and Li, Peipei and Xia, Shuhan and Cui, Xing and Huang, Linzhi and Huang, Huaibo and Deng, Weihong and He, Zhaofeng},
  journal={arXiv preprint arXiv:2406.08772},
  year={2024}
}

@article{xuan2024lemma,
  title={LEMMA: Towards LVLM-Enhanced Multimodal Misinformation Detection with External Knowledge Augmentation},
  author={Xuan, Keyang and Yi, Li and Yang, Fan and Wu, Ruochen and Fung, Yi R and Ji, Heng},
  journal={arXiv preprint arXiv:2402.11943},
  year={2024}
}

@article{DeepfakeSurvey2024Wang,
author = {Wang, Tianyi and Liao, Xin and Chow, Kam Pui and Lin, Xiaodong and Wang, Yinglong},
title = {Deepfake Detection: A Comprehensive Survey from the Reliability Perspective},
year = {2024},
volume = {57},
number = {3},
issn = {0360-0300},
doi = {10.1145/3699710},
journal = {ACM Comput. Surv.},
month = nov,
articleno = {58},
numpages = {35}
}

@article{NoiseDF2023Wang, 
title={Noise Based Deepfake Detection via Multi-Head Relative-Interaction}, 
author={Wang, Tianyi and Chow, Kam Pui}, 
journal={Proceedings of the AAAI Conference on Artificial Intelligence}, 
year={2023}, 
volume={37}, 
DOI={10.1609/aaai.v37i12.26701}, 
number={12}, 
month={Jun.}, 
pages={14548-14556} 
}

@article{berman2024aimisinformation,
  author  = {Will Henshall},
  title   = {Tech Companies Are Taking Action on AI Election Misinformation. Will It Matter?},
  journal = {Time},
  year    = {2023},
  url     = {https://time.com/6333288/tech-companies-ai-misinformation/},
  note    = {Accessed: 2025-04-11}
}

@article{borges2022infodemics,
  author    = {Borges do Nascimento, Igor J. and Pizarro, Anna B. and Almeida, Jo{\~a}o M. and Azzopardi-Muscat, Natasha and Gon{\c{c}}alves, Miguel A. and Bj{\"o}rklund, Maria and Novillo-Ortiz, Douglas},
  title     = {Infodemics and health misinformation: a systematic review of reviews},
  journal   = {Bulletin of the World Health Organization},
  volume    = {100},
  number    = {9},
  pages     = {544--561},
  year      = {2022},
  month     = sep,
  doi       = {10.2471/BLT.21.287654},
  pmid      = {36062247},
  pmcid     = {PMC9421549},
  note      = {Epub 2022 Jun 30}
}

@article{wsj2025falseTariffHeadline,
  title   = {The False Tariff Headline That Sent Stocks on a \$2 Trillion Ride},
  journal = {The Wall Street Journal},
  year    = {2025},
  month   = apr,
  day     = {8},
  url     = {https://www.wsj.com/finance/stocks/the-false-tariff-headline-that-sent-stocks-on-a-2-trillion-ride-2224ef75},
  note    = {Accessed: 2025-04-11}
}

@article{ARCURI2023106130,
title = {Does fake news impact stock returns? Evidence from US and EU stock markets},
journal = {Journal of Economics and Business},
volume = {125-126},
pages = {106130},
year = {2023},
issn = {0148-6195},
doi = {https://doi.org/10.1016/j.jeconbus.2023.106130},
url = {https://www.sciencedirect.com/science/article/pii/S0148619523000231},
author = {Maria Cristina Arcuri and Gino Gandolfi and Ivan Russo}
}

@ARTICLE{11010889,
  author={Zou, Mian and Yu, Baosheng and Zhan, Yibing and Lyu, Siwei and Ma, Kede},
  journal={IEEE Transactions on Circuits and Systems for Video Technology}, 
  title={Semantics-Oriented Multitask Learning for DeepFake Detection: A Joint Embedding Approach}, 
  year={2025},
  volume={35},
  number={10},
  pages={9950-9963},
  doi={10.1109/TCSVT.2025.3572508}}

@article{reuters2025duterteDisinformation,
  title   = {Fake accounts drove praise of Duterte and now target Philippine election},
  journal = {Reuters},
  year    = {2025},
  month   = apr,
  day     = {11},
  url     = {https://www.reuters.com/world/asia-pacific/fake-accounts-drove-praise-duterte-now-target-philippine-election-2025-04-11/},
  note    = {Accessed: 2025-04-11}
}

@ARTICLE{9408664,
  author={Hu, Juan and Liao, Xin and Wang, Wei and Qin, Zheng},
  journal={IEEE Transactions on Circuits and Systems for Video Technology}, 
  title={Detecting Compressed Deepfake Videos in Social Networks Using Frame-Temporality Two-Stream Convolutional Network}, 
  year={2022},
  volume={32},
  number={3},
  pages={1089-1102},
  doi={10.1109/TCSVT.2021.3074259}}

@ARTICLE{10487975,
  author={Qiao, Tong and Shao, Hang and Xie, Shichuang and Shi, Ran},
  journal={IEEE Transactions on Circuits and Systems for Video Technology}, 
  title={Unsupervised Generative Fake Image Detector}, 
  year={2024},
  volume={34},
  number={9},
  pages={8442-8455},
  doi={10.1109/TCSVT.2024.3383833}}

@inproceedings{wang2023dire,
  title={Dire for diffusion-generated image detection},
  author={Wang, Zhendong and Bao, Jianmin and Zhou, Wengang and Wang, Weilun and Hu, Hezhen and Chen, Hong and Li, Houqiang},
  booktitle={Proceedings of the IEEE/CVF International Conference on Computer Vision},
  pages={22445--22455},
  year={2023}
}

@inproceedings{yu2024mm,
  title={Mm-vet: Evaluating large multimodal models for integrated capabilities},
  author={Yu, Weihao and Yang, Zhengyuan and Li, Linjie and Wang, Jianfeng and Lin, Kevin and Liu, Zicheng and Wang, Xinchao and Wang, Lijuan},
  booktitle={International conference on machine learning},
  year={2024},
  organization={PMLR}
}

@article{hendryckstest2021,
  title={Measuring Massive Multitask Language Understanding},
  author={Dan Hendrycks and Collin Burns and Steven Basart and Andy Zou and Mantas Mazeika and Dawn Song and Jacob Steinhardt},
  journal={Proceedings of the International Conference on Learning Representations (ICLR)},
  year={2021}
}

@inproceedings{qi2024sniffer,
  title={Sniffer: Multimodal large language model for explainable out-of-context misinformation detection},
  author={Qi, Peng and Yan, Zehong and Hsu, Wynne and Lee, Mong Li},
  booktitle={Proceedings of the IEEE/CVF conference on computer vision and pattern recognition},
  pages={13052--13062},
  year={2024}
}

@inproceedings{muller2020multimodal,
  title={Multimodal analytics for real-world news using measures of cross-modal entity consistency},
  author={M{\"u}ller-Budack, Eric and Theiner, Jonas and Diering, Sebastian and Idahl, Maximilian and Ewerth, Ralph},
  booktitle={Proceedings of the 2020 international conference on multimedia retrieval},
  pages={16--25},
  year={2020}
}

@inproceedings{aneja2023cosmos,
  title={COSMOS: catching out-of-context image misuse using self-supervised learning},
  author={Aneja, Shivangi and Bregler, Chris and Nie{\ss}ner, Matthias},
  booktitle={Proceedings of the AAAI conference on artificial intelligence},
  volume={37},
  number={12},
  pages={14084--14092},
  year={2023}
}

@inproceedings{abdelnabi2022open,
  title={Open-domain, content-based, multi-modal fact-checking of out-of-context images via online resources},
  author={Abdelnabi, Sahar and Hasan, Rakibul and Fritz, Mario},
  booktitle={Proceedings of the IEEE/CVF conference on computer vision and pattern recognition},
  pages={14940--14949},
  year={2022}
}

@article{luo2021newsclippings,
  title={Newsclippings: Automatic generation of out-of-context multimodal media},
  author={Luo, Grace and Darrell, Trevor and Rohrbach, Anna},
  journal={arXiv preprint arXiv:2104.05893},
  year={2021}
}

@inproceedings{jaiswal2017multimedia,
  title={Multimedia semantic integrity assessment using joint embedding of images and text},
  author={Jaiswal, Ayush and Sabir, Ekraam and AbdAlmageed, Wael and Natarajan, Premkumar},
  booktitle={Proceedings of the 25th ACM international conference on Multimedia},
  pages={1465--1471},
  year={2017}
}

@inproceedings{papadopoulos2023synthetic,
  title={Synthetic misinformers: Generating and combating multimodal misinformation},
  author={Papadopoulos, Stefanos-Iordanis and Koutlis, Christos and Papadopoulos, Symeon and Petrantonakis, Panagiotis},
  booktitle={Proceedings of the 2nd ACM International Workshop on Multimedia AI against Disinformation},
  pages={36--44},
  year={2023}
}

@article{schlichtkrull2023averitec,
  title={Averitec: A dataset for real-world claim verification with evidence from the web},
  author={Schlichtkrull, Michael and Guo, Zhijiang and Vlachos, Andreas},
  journal={Advances in Neural Information Processing Systems},
  volume={36},
  pages={65128--65167},
  year={2023}
}

@ARTICLE{10290956,
  author={Liu, Miao and Wang, Jing and Qian, Xinyuan and Li, Haizhou},
  journal={IEEE Transactions on Circuits and Systems for Video Technology}, 
  title={Audio-Visual Temporal Forgery Detection Using Embedding-Level Fusion and Multi-Dimensional Contrastive Loss}, 
  year={2024},
  volume={34},
  number={8},
  pages={6937-6948},
  doi={10.1109/TCSVT.2023.3326694}}

@inproceedings{jiang2020hover,
  title={{HoVer}: A Dataset for Many-Hop Fact Extraction And Claim Verification},
  author={Yichen Jiang and Shikha Bordia and Zheng Zhong and Charles Dognin and Maneesh Singh and Mohit Bansal.},
  booktitle={Findings of the Conference on Empirical Methods in Natural Language Processing ({EMNLP})},
  year={2020}
}

@inproceedings{aly2021fact,
  title={The fact extraction and VERification over unstructured and structured information (FEVEROUS) shared task},
  author={Aly, Rami and Guo, Zhijiang and Schlichtkrull, Michael and Thorne, James and Vlachos, Andreas and Christodoulopoulos, Christos and Cocarascu, Oana and Mittal, Arpit},
  booktitle={Proceedings of the Fourth Workshop on Fact Extraction and VERification (FEVER)},
  pages={1--13},
  year={2021}
}
\begin{IEEEbiography}[{\raisebox{0.4\height}{\includegraphics[width=1in,height=1.25in,clip,keepaspectratio]{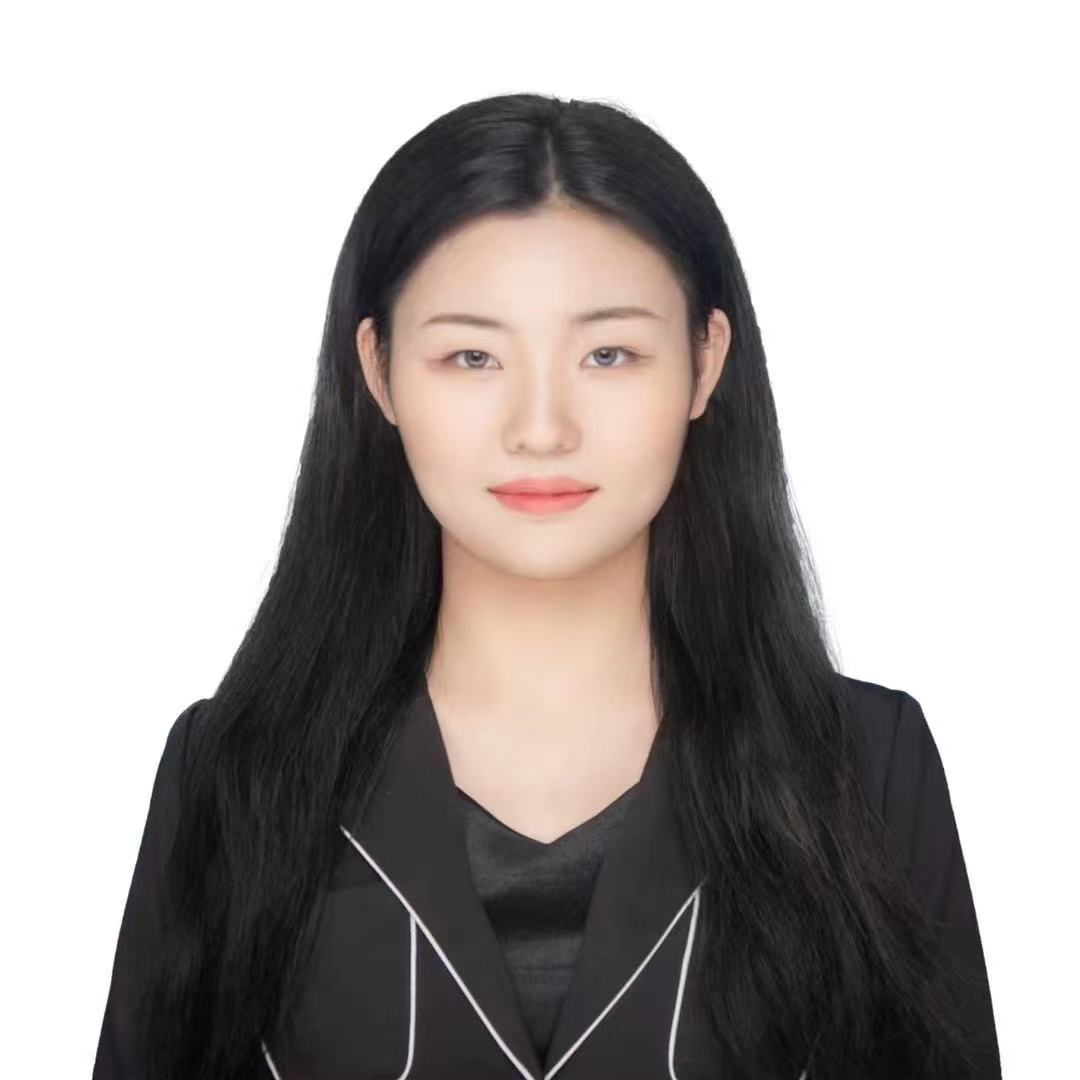}}}]
{DANNI XU} is currently a Ph.D. student with the
School of Computing, National University of Singapore. She received the B.S. degree in computer science from Wuhan University, Wuhan,
China, in 2019, and the M.S. degree in computer science from the same
university in 2022. Her research interests include misinformation detection, multimodal learning, and human behavior analysis.
% She has authored or coauthored papers in top conferences, including ACM MM, IJCAI, CVPR, and AAAI. 
\end{IEEEbiography}

\begin{IEEEbiography}[{\raisebox{0.0\height}{\includegraphics[width=1in,height=1.25in,clip,keepaspectratio]{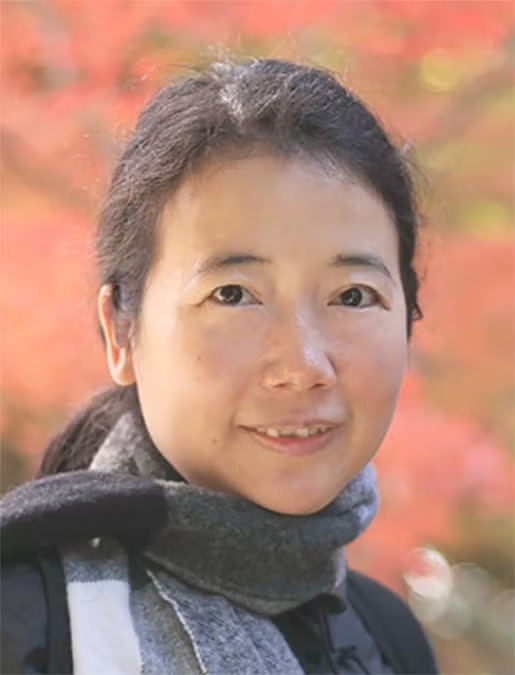}}}]
{SHAOJING FAN} is currently a Senior Lecturer in the Department of Electrical and Computer Engineering (ECE), National University of Singapore (NUS). Before joining ECE, she was a Senior Research Fellow in the School of Computing at NUS. Prior to that, she worked as a Senior Research Engineer at the Institute for Infocomm Research (I2R), part of Singapore’s Agency for Science, Technology, and Research (A*STAR).
Dr. Fan received her B.E. and M.E. degrees from South China University of Technology in 2001 and 2004, respectively. She earned her D.Phil. from Ningbo University, with the majority of her research conducted at the I2R, A*STAR, between 2010 and 2015. 
Dr. Fan’s research interests include computer vision, cognitive vision, computational social science, and experimental psychology. Her work has been published in leading international journals and conferences, including TPAMI, CVPR, AAAI, and SIGGRAPH Asia. She is a member of the Association for Computing Machinery (ACM) and serves as a reviewer for several top-tier journals and conferences, such as TPAMI, CVPR, AAAI, ICLR, ICML, and NeurIPS.
\end{IEEEbiography}

\begin{IEEEbiography}[{\raisebox{0.25\height}{\includegraphics[width=1in,height=1.25in,clip,keepaspectratio]{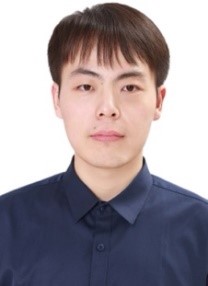}}}]{HARRY CHENG} is a Research Fellow at the National University of Singapore, Singapore. He received his Ph.D. degree from Shandong University, China. He has authored or coauthored several papers in top conferences and journals, including NeurIPS, ICCV, ACM MM, and IEEE TMM, and he serves as a regular reviewer for several journals such as IEEE TPAMI, IEEE TIP, IEEE TIFS, IEEE TKDE, IEEE TMM, and ACM ToMM.
\end{IEEEbiography}

\begin{IEEEbiography}[{\raisebox{0.\height}{\includegraphics[width=1in,height=1.25in,clip,keepaspectratio]{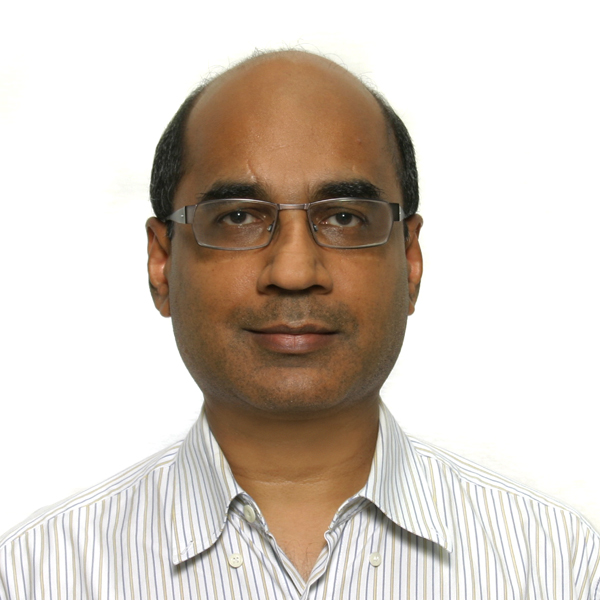}}}]{MOHAN KANKANHALLI} (Fellow, IEEE) is Provost's Chair Professor of Computer Science at the National University of Singapore (NUS) and the Deputy Executive Chairman of AI Singapore. He is also the Director of NUS AI Institute, where he leads initiatives on multimodal models and trustworthy machine learning. Mohan obtained his BTech from IIT Kharagpur and MS \& PhD from the Rensselaer Polytechnic Institute. Mohan’s research interests are in Multimodal Computing, Computer Vision and Trustworthy AI. Mohan was a member of World Economic Forum's 2023-2024 Global Future Council on the Future of Artificial Intelligence. He is a member of ACM’s Global Technology Policy Council. Mohan is a Fellow of IEEE, IAPR and ACM.
\end{IEEEbiography}

%%
%% If your work has an appendix, this is the place to put it.
\clearpage
\appendices
\section*{Appendix}
\addcontentsline{toc}{section}{Appendix}
\renewcommand{\thefigure}{S\arabic{figure}}
\setcounter{figure}{0}
\renewcommand{\thetable}{S\arabic{table}}
\setcounter{table}{0}
\setcounter{page}{1}
\subsection{Data Filtering and Leakage Control}
\label{app:dataset}
Table~\ref{tab:ocr_watermark_prompt} and Table~\ref{tab:image_verification_prompt} present the prompts used to filter samples in the RW-Post dataset, aiming to reduce label leakage from images and to exclude samples where only the text contributes to veracity assessment.
\begin{table}[H]
\centering
\footnotesize
\setlength{\tabcolsep}{4pt}
\renewcommand{\arraystretch}{0.95}

\begin{tcolorbox}[
colback=gray!3,
colframe=gray!40,
title={Prompt for Image Verification: Label-Watermark Detection},
boxsep=2pt,
left=3pt,right=3pt,top=3pt,bottom=3pt
]

\textbf{Role:} You are an expert in image verification and multimodal misinformation analysis.

\vspace{2pt}
\textbf{Task:} Given an input image, extract any textual content via OCR and determine whether the image contains watermark-like labels commonly used by fact-checking organizations (e.g., “fake”, “altered”, “misleading”, “satire”, etc.).

\vspace{2pt}
\textbf{Requirements:}
\begin{enumerate}
\item \textbf{Extract OCR text} from the image as accurately as possible.
\item \textbf{Detect watermark-like keywords} indicating the image has been labeled or classified.  
\textit{fake, false, altered, misleading, miscaptioned, scam, satire, outdated, unproven, mixture, correct attribution, mostly false, mostly true, misattributed, composite image, no evidence, April Fools’ Day}.
\item \textbf{Provide a concise explanation} for your decision.
\end{enumerate}

\vspace{2pt}
\textbf{Output Format:}
\begin{lstlisting}[language=json]
{
  "watermarked": true/false,
  "matched_keywords": ["..."],
  "reason": ""
}
\end{lstlisting}

\end{tcolorbox}
\caption{Prompt for evaluating whether an image contains label-leaked watermarks}
\label{tab:ocr_watermark_prompt}
\end{table}

\begin{table}[H]
\centering
\footnotesize
\renewcommand{\arraystretch}{0.95}

\begin{tcolorbox}[
colback=gray!3,
colframe=gray!40,
title={Prompt for Image Necessity Assessment},
boxsep=2pt,
left=3pt,right=3pt,top=3pt,bottom=3pt
]

\textbf{Role:} You are a fact-checking expert.

\vspace{2pt}
\textbf{Task:} Determine whether the claim \textbf{requires an image} for verification.

\vspace{2pt}
\textbf{Criteria for when an image is considered  ``required'':}
\begin{itemize}
    \item The image would serve as direct evidence for the claim (e.g., confirming an event);
    \item The image would significantly influence users’ judgment of the claim’s truthfulness.
\end{itemize}

\vspace{2pt}
\textbf{Output:}
\begin{enumerate}
\item \texttt{yes} / \texttt{no} / \texttt{uncertain}
\item Brief explanation
\end{enumerate}

\vspace{2pt}
\textbf{Format:}
\begin{lstlisting}[language=json]
{
  "image_required": "yes | no | unsure",
  "reason": ""
}
\end{lstlisting}

\end{tcolorbox}
\caption{Prompt for evaluating whether image context is required for a claim}
\label{tab:image_verification_prompt}
\end{table}

\subsection{Prompts of Agents in AgentFact}
\label{app:agent}
\noindent
\begin{minipage}{\linewidth}
\footnotesize

\begin{tcolorbox}[
colback=gray!3,
colframe=gray!40,
title={Prompt for Strategy Planning Agent (Agent-SP)},
boxsep=2pt,
left=3pt,right=3pt,top=3pt,bottom=3pt
]

\textbf{Role:} You are a fact-checking plan designer in a multi-agent fact-checking framework.

\vspace{2pt}
\textbf{Task:} Given the post content, claim, and context, generate or refine a verification plan that guides efficient and accurate fact checking. Apply appropriate fact-checking techniques (e.g., Divide and Conquer, Origin Tracing, Chain of Evidence, Cross-Verification, Temporal Consistency, Source Credibility, Logical Consistency).

\vspace{2pt}
\textbf{Your Output Must Contain:}
\begin{enumerate}
\item \textbf{Validation Logic}  
A concise, structured reasoning plan for analyzing the claim, indicating which techniques apply.

\item \textbf{Validation List (up to 3 items)}  
Original sentences from the post requiring verification. Each sentence must explicitly contain all essential information (no pronouns). Leave empty if none require direct validation.

\item \textbf{Search List (up to 3 items)}  
Key information that must be externally retrieved for fact checking. Ordered by priority and non-overlapping with the validation list.
\end{enumerate}

\vspace{2pt}
\textbf{Constraints:}
\begin{itemize}
\item The validation list and search list must not overlap or contain redundant items.  
\item Include only information truly necessary for fact checking.  
\item Keep the plan concise but complete.
\end{itemize}

\vspace{2pt}
\textbf{Output Format:}
\begin{lstlisting}[language=json]
{
  "reasoning_steps": [ {"step": "", "method": "", "details": ""}, ... ],
  "validation_list": [ {"sentence": "", "explanation": ""}, ... ],
  "search_list": [ {"information_needed": ""}, ... ]
}
\end{lstlisting}

\end{tcolorbox}
\end{minipage}

\vspace{6pt}

\noindent
\begin{minipage}{\linewidth}
\footnotesize

\begin{tcolorbox}[
colback=gray!3,
colframe=gray!40,
title={Prompt for Text Evidence Retrieval and Validation Agent (Agent-TR-1): Query Generation},
boxsep=2pt,
left=3pt,right=3pt,top=3pt,bottom=3pt
]

\textbf{Role:} You are a fact-checking retrieval assistant responsible for generating high-quality search queries.

\vspace{2pt}
\textbf{Task:} Given an information need, claim and post context, produce a small set of SEO-effective, high-intent search queries suitable for Google. Queries should be specific, long-tail, and directly usable for evidence retrieval. Avoid redundancy with previous queries and ensure each query targets distinct information.

\vspace{2pt}
\textbf{Guidelines:}
\begin{itemize}
\item Generate focused, high-intent long-tail queries.  
\item Avoid duplicate or semantically similar queries.  
\item When necessary, break complex information into smaller searchable components.  
\item Tailor queries for reliability and relevance.  
\item Keep queries concise and avoid vague filler phrases.
\end{itemize}

\vspace{2pt}
\textbf{Output Requirements:}
\begin{itemize}
\item Generate at most one query per information item.  
\item Include only information worth retrieving externally.  
\item Queries must contain explicit keywords (no pronouns).
\end{itemize}

\vspace{2pt}
\textbf{Output Format:}
\begin{lstlisting}[language=json]
{ "queries": [ "best search query 1", "best search query 2" ] }
\end{lstlisting}

\end{tcolorbox}
\end{minipage}

\clearpage

\noindent
\begin{minipage}{\linewidth}
\footnotesize

\begin{tcolorbox}[
colback=gray!3,
colframe=gray!40,
title={Prompt for Text Evidence Retrieval and Validation Agent (Agent-TR-2): Source Reliability Assessment},
boxsep=2pt,
left=3pt,right=3pt,top=3pt,bottom=3pt
]

\textbf{Role:} You are an expert in digital literacy and online misinformation analysis. Your task is to assess the reliability and intent of a given website domain.

\vspace{2pt}
\textbf{Task:} Given a URL or domain, classify the source into one of the categories: \texttt{reliable}, \texttt{unreliable}, \texttt{satire}, \texttt{unsure}, \texttt{factcheck}.

\vspace{2pt}
\textbf{Requirements:}
\begin{enumerate}
\item \textbf{Identify the domain} (e.g., \texttt{cnn.com}, \texttt{theonion.com}).  
\item \textbf{Evaluate source characteristics}, including whether it is:  
\begin{itemize}[leftmargin=*,itemsep=1pt,topsep=1pt]
\item listed in misinformation / disinformation databases;  
\item frequently debunked by reputable fact-checkers;  
\item known satire or parody;  
\item legitimate, professional journalism;  
\item a fact-checking organization.  
\end{itemize}
\item \textbf{Provide a concise classification explanation} (2–4 sentences).  
\item \textbf{Describe how a fact-checker should treat information} from this domain:  
\begin{itemize}[leftmargin=*,itemsep=1pt,topsep=1pt]
\item \textbf{Positive use}: information can generally be trusted;  
\item \textbf{Reverse use}: presence of information is itself evidence of low credibility;  
\item \textbf{Neutral/unsure}: requires strong corroboration.  
\end{itemize}
\end{enumerate}

\vspace{2pt}
\textbf{Output Format:}
\begin{lstlisting}[language=json]
{
  "source_identification": "",
  "type": "<reliable | unreliable | satire | unsure | factcheck>",
  "reasoning": "",
  "fact_checker_usage": ""
}
\end{lstlisting}

\end{tcolorbox}
\end{minipage}

\vspace{6pt}

\noindent
\begin{minipage}{\linewidth}
\footnotesize

\begin{tcolorbox}[
colback=gray!3,
colframe=gray!40,
title={Prompt for Reasoning Agent (Agent-R)},
boxsep=2pt,
left=3pt,right=3pt,top=3pt,bottom=3pt
]

\textbf{Role:} You are an expert fact-checking reasoning agent. Your task is to analyze the claim using structured reasoning steps and evidence with source-reliability judgments.

\vspace{2pt}
\textbf{Task:} Given the claim, post context, reasoning plan, retrieved text evidence (annotated with source reliability), and image-analysis results, execute each reasoning step in order and identify which evidence is relevant, irrelevant, or not required.

\vspace{2pt}
\textbf{Requirements:}
\begin{enumerate}

\item \textbf{Interpret the claim.}  
Produce a concise paraphrase capturing the core factual assertion. Output as \texttt{"my\_understanding\_of\_claim"}.

\item \textbf{Follow the reasoning plan strictly.}  
Execute each step in the provided sequence.

\item \textbf{Evaluate evidence step-by-step.}  
For each reasoning step:
\begin{itemize}[leftmargin=*,itemsep=1pt,topsep=1pt]
\item Identify relevant evidence and explain how it supports the analysis.  
\item If no evidence is needed, state \texttt{"Evidence not required"}.  
\item If no relevant evidence exists, state \texttt{"Relevant evidence not found"}.  
\item You may optionally provide reliable evidence based on your own knowledge (with source, link, and reputation).
\end{itemize}

\item \textbf{Restrictions.}
\begin{itemize}[leftmargin=*,itemsep=1pt,topsep=1pt]
\item The post itself cannot be used as evidence.  
\item Conflicting evidence must be evaluated with respect to source reliability.  
\item Absence of evidence does not imply falsity—assign confidence cautiously.
\end{itemize}

\item \textbf{Final confidence score (1–5).}  
Score reflects the sufficiency and reliability of evidence supporting the final assessment.

\end{enumerate}

\vspace{2pt}
\textbf{Output Format:}
\begin{lstlisting}[language=json]
{
  "my_understanding_of_claim": "",
  "validation_result": {
    "reasoning_steps": [
      {
        "step_name": "",
        "description": "",
        "analysis_result": "",
        "relevant_evidence_summary": "",
        "relevant_text_evidence_list": [],
        "relevant_image_evidence_list": [],
        "evidence_based_on_my_knowledge": []
      }
    ],
    "direct_fact_check_evidence": {
      "analysis_result": "",
      "relevant_evidence_summary": "",
      "relevant_text_evidence_list": []
    },
    "final_sufficiency_confidence": ""
  }
}
\end{lstlisting}

\vspace{2pt}
\textbf{Note:} Keep reasoning concise but evidence-grounded.

\end{tcolorbox}
\end{minipage}

\clearpage

\noindent
\begin{minipage}{\linewidth}
\footnotesize

\begin{tcolorbox}[
colback=gray!3,
colframe=gray!40,
title={Prompt for Image Retrieval and Analysis Agent (IR-1): Image Matching and Manipulation Detection},
boxsep=2pt,
left=3pt,right=3pt,top=3pt,bottom=3pt
]

\textbf{Role:} You are an image comparison assistant supporting multimodal fact checking.

\vspace{2pt}
\textbf{Task:} Given a post image and a retrieved evidence image, analyze their visual relationship and assess whether the post image shows signs of manipulation.

\vspace{2pt}
\textbf{Step 1: Classify Image Relationship}
Determine which of the following categories best describes the relationship:
\begin{itemize}
\item \textbf{Potentially From Same Source}: Nearly identical composition, contents, and configuration.  
\item \textbf{Same Event, Different Content}: Depict the same real-world event but differ in angle, timing, or framing.  
\item \textbf{No Close Relationship}: Unrelated subjects, events, or contexts.
\end{itemize}

\vspace{2pt}
\textbf{Step 2: Manipulation Assessment}  
If the relationship is not \textit{No Close Relationship}, evaluate whether the post image shows signs of tampering based on:
\begin{itemize}
\item Self-analysis of the post image  
\item Direct comparison with the evidence image  
\end{itemize}

\vspace{2pt}
\textbf{Output Requirements:}
Provide a concise explanation for the relationship classification, estimate tampering probability (0–100), and give a short reasoning summary.

\vspace{2pt}
\textbf{Output Format:}
\begin{lstlisting}[language=json]
{
  "relationship": "",
  "relationship_reasoning": "",
  "tampering_probability": "",
  "tampering_reasoning": "",
  "confidence": ""
}
\end{lstlisting}

\vspace{2pt}
\textbf{Notes:}
\begin{itemize}
\item Focus on visually discriminative and fact-check–relevant features.  
\item Keep explanations clear and grounded in observable visual evidence.  
\item Leave tampering fields empty if the relationship is "No Close Relationship".
\end{itemize}

\end{tcolorbox}
\end{minipage}

\vspace{6pt}

\noindent
\begin{minipage}{\linewidth}
\footnotesize

\begin{tcolorbox}[
colback=gray!3,
colframe=gray!40,
title={Prompt for Image Retrieval and Analysis Agent (IR-2): Image Miscaption Detection},
boxsep=2pt,
left=3pt,right=3pt,top=3pt,bottom=3pt
]

\textbf{Role:} You are an image–text consistency analysis assistant for fact checking.

\vspace{2pt}
\textbf{Task:} Given a post image, its claim, and a text context of an evidence image, determine whether the post image is miscaptioned.

\vspace{2pt}
\textbf{Step 1: Interpret the Claim}  
Provide a concise paraphrase capturing the core factual assertion of the claim. Output as \texttt{"my\_understanding\_of\_claim"}.

\vspace{2pt}
\textbf{Step 2: Understand the Evidence}  
Summarize the evidence image and text in your own words, including its event, context, or purpose. Assess alignment with the claim.

\vspace{2pt}
\textbf{Step 3: Compare Temporal and Contextual Information}  
Compare dates, locations, or actors if available and identify discrepancies.

\vspace{2pt}
\textbf{Step 4: Miscaption Analysis}  
\begin{itemize}
\item comes from an unrelated event, place, or time;  
\item misrepresents who is involved or what is happening;  
\item substantially distorts the factual context.
\end{itemize}

\vspace{2pt}
\textbf{Output Format:}
\begin{lstlisting}[language=json]
{
  "my_understanding_of_claim": "",
  "Miscaption Rate": "",
  "Reasoning": ""
}
\end{lstlisting}

\vspace{2pt}
\textbf{Scoring Guide}
\begin{itemize}
\item 0–20: Image accurately supports the claim  
\item 30–50: Generally aligned but missing context  
\item 60–80: Provides a misleading impression  
\item 90–100: Unrelated or strongly contradicts the claim  
\end{itemize}

\vspace{2pt}
\textbf{Note:} Focus strictly on factual alignment between the claim and image.

\end{tcolorbox}
\end{minipage}

\vspace{6pt}

\noindent
\begin{minipage}{\linewidth}
\footnotesize

\begin{tcolorbox}[
colback=gray!3,
colframe=gray!40,
title={Prompt for Explanation Generation Agent (Agent-EG)},
boxsep=2pt,
left=3pt,right=3pt,top=3pt,bottom=3pt
]

\textbf{Role:} You are the final reasoning and explanation agent in a fact-checking framework.

\vspace{2pt}
\textbf{Task:} Given the claim, post context, textual evidence, image-analysis results, and previous reasoning outputs, produce a final authenticity assessment with explanation and confidence.

\vspace{2pt}
\textbf{Requirements:}
\begin{enumerate}

\item \textbf{Interpret the claim.}  
Provide a concise paraphrase.

\item \textbf{Assess claim authenticity.}
\begin{itemize}[leftmargin=*,itemsep=1pt,topsep=1pt]
\item \textbf{Coarse label}: TRUE / FALSE  
\item \textbf{Fine-grained label}: TRUE / FALSE / UNPROVEN  
\end{itemize}

\item \textbf{Decision principles.}
\begin{itemize}[leftmargin=*,itemsep=1pt,topsep=1pt]
\item Strong refuting evidence → FALSE  
\item Strong supporting evidence → TRUE  
\item Insufficient evidence → UNPROVEN  
\end{itemize}

\item \textbf{Reasoning and evidence citation.}
Cite text and image evidence IDs.

\item \textbf{Confidence score (1–5).}

\end{enumerate}

\vspace{2pt}
\textbf{Output Format:}
\begin{lstlisting}[language=json]
{
  "my_understanding_of_claim": "",
  "validation_result": {
    "2-class_authenticity_label": "",
    "3-class_authenticity_label": "",
    "reasoning_logic": "",
    "key_points": [
      "1. ...",
      "2. ...",
      "3. ..."
    ]
  },
  "confidence_level": ""
}
\end{lstlisting}

\vspace{2pt}
\textbf{Note:}
\begin{itemize}
\item Focus on the truth of the claim itself.  
\item Emphasize factual alignment and evidence reliability.
\end{itemize}

\end{tcolorbox}
\end{minipage}

\subsection{Experimental Details: Fact-Checking Domain Filtering}
\label{filter_domains}
To mitigate potential \textit{label leakage} during evidence retrieval—particularly when external search engines return fact-checking articles that explicitly state the veracity of the claim—we applied a domain filtering strategy.

We excluded any evidence items whose source URLs contained substrings associated with known fact-checking organizations or services. This ensures that the model cannot trivially infer labels by relying on already-verified ground truth statements from professional sources.
The filtering was implemented by checking whether the domain name or URL of the retrieved evidence contains any of the following substrings, which are associated with known fact-checking organizations or services.
For brevity, a representative list is shown below; the complete list is available in our code repository.
\begin{multicols}{3}
\begin{itemize}
  \item snopes
  \item politifact
  \item factcheck
  \item truthorfiction
  \item hoax-slayer
  \item eadstories
  \item opensecrets
  \item fullfact
  \item checkyourfact
  \item realitycheck
  \item fact-check
  \item ...
\end{itemize}
\end{multicols}

This filter was applied consistently to all models that utilized external search (e.g., LEMMA, DEFAME, AgentFact) during evidence retrieval.
By removing these high-authority verification sources, we ensure that the model’s performance more accurately reflects its ability to reason over raw evidence, rather than memorize or match against curated labels.

\subsection{Additional Case Study}
\begin{figure}[H]
    \centering

    % ====================
    % Top: Image | Claim (Left–Right)
    % ====================
    \begin{minipage}[c]{0.36\linewidth}
        \centering
        \includegraphics[width=\linewidth]{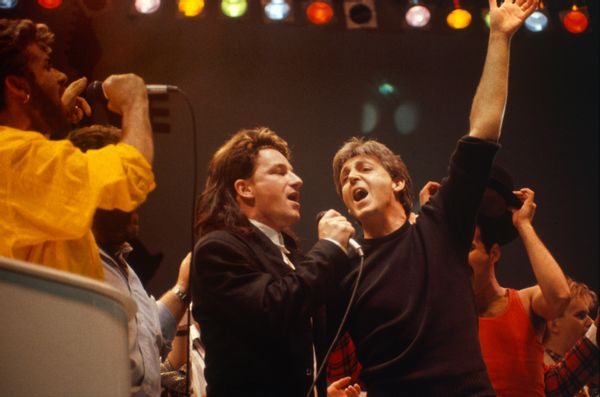}
    \end{minipage}
    \begin{minipage}[c]{0.6\linewidth}
        \footnotesize
        \textbf{\textsc{Claim:}}\\
        A photograph genuinely shows singers Bono, Paul McCartney, and
        Freddie Mercury performing together at the Live Aid benefit
        concert in 1985.
    \end{minipage}

    \vspace{0.3em}
\setlength{\tabcolsep}{4pt}
\renewcommand{\arraystretch}{1.15}
\small
\begin{tabularx}{\linewidth}{X}
\toprule

\textsc{Label} \quad \quad \quad \quad \quad  
\textbf{Model:} \textsc{False} \quad
\textbf{Human:} \textsc{True} \\

\midrule
\textsc{Reasoning} \\
\textbf{Model:}
Although Bono, Paul McCartney, and Freddie Mercury all appeared at
the Live Aid concert on July 13, 1985,
there is no explicit evidence confirming that the three
artists performed together on the same stage.\\[2pt]
\textbf{Human:}
The photograph is authentic and originates from the 1985 Live Aid
concert. Multimodal evidence confirms that Bono, Paul McCartney,
and Freddie Mercury appeared together during the same on-stage
segment. \\

\midrule
\textsc{Key Points} \\
\textbf{Model:}
(1) All three artists \hlred{individually} participated in Live Aid;
(2) no official report explicitly documents a joint performance.\\[2pt]
\textbf{Human:}
(1) The image is verified as originating from Live Aid (1985);
(2) video footage captures \hlgreen{on-stage interaction} between Bono and Freddie Mercury;
(3) archival photographs show Paul McCartney appearing alongside them. \\

\midrule
\textsc{Evidence} \\
\textbf{Model Evidence:}
[1] Text from the-paulmccartney-project.com:
Paul McCartney performed at Live Aid on July 13, 1985. 
{}[2] Text from rollingstone.com: U2 (with Bono) performed at Live Aid on July 13, 1985.\\
\textbf{Human Evidence:}
[1] Video from YouTube:
Bono and Freddie Mercury \hlgreen{singing into the same microphone} on stage. 
{}[2] Image/Text from Getty Images:
Photographs depicting Bono, Paul McCartney, and Freddie Mercury
appearing together during the concert. \\

\midrule
\textsc{Comparison} \\
\textbf{The model rejects the claim due to the absence of explicit textual
documentation for a three-person performance, whereas the ground
truth confirms the claim by integrating complementary multimodal
evidence that correctly attributes the photograph to the Live Aid
concert scene.} \\

\bottomrule
\end{tabularx}

\caption{
Case study of an incorrectly classified claim with weak contextual grounding.
}
\label{fig:live_aid_vertical_case}
\end{figure}

Fig.~\ref{fig:live_aid_vertical_case} shows an additional failure case where the model relies on textual evidence but fails to incorporate multimodal signals, leading to incorrect prediction. 
This highlights limitations in contextual grounding.

\subsection{Human evaluation}
\label{annotation_interface}
\begin{figure}[h]
    \centering
    \includegraphics[width=0.8\linewidth]{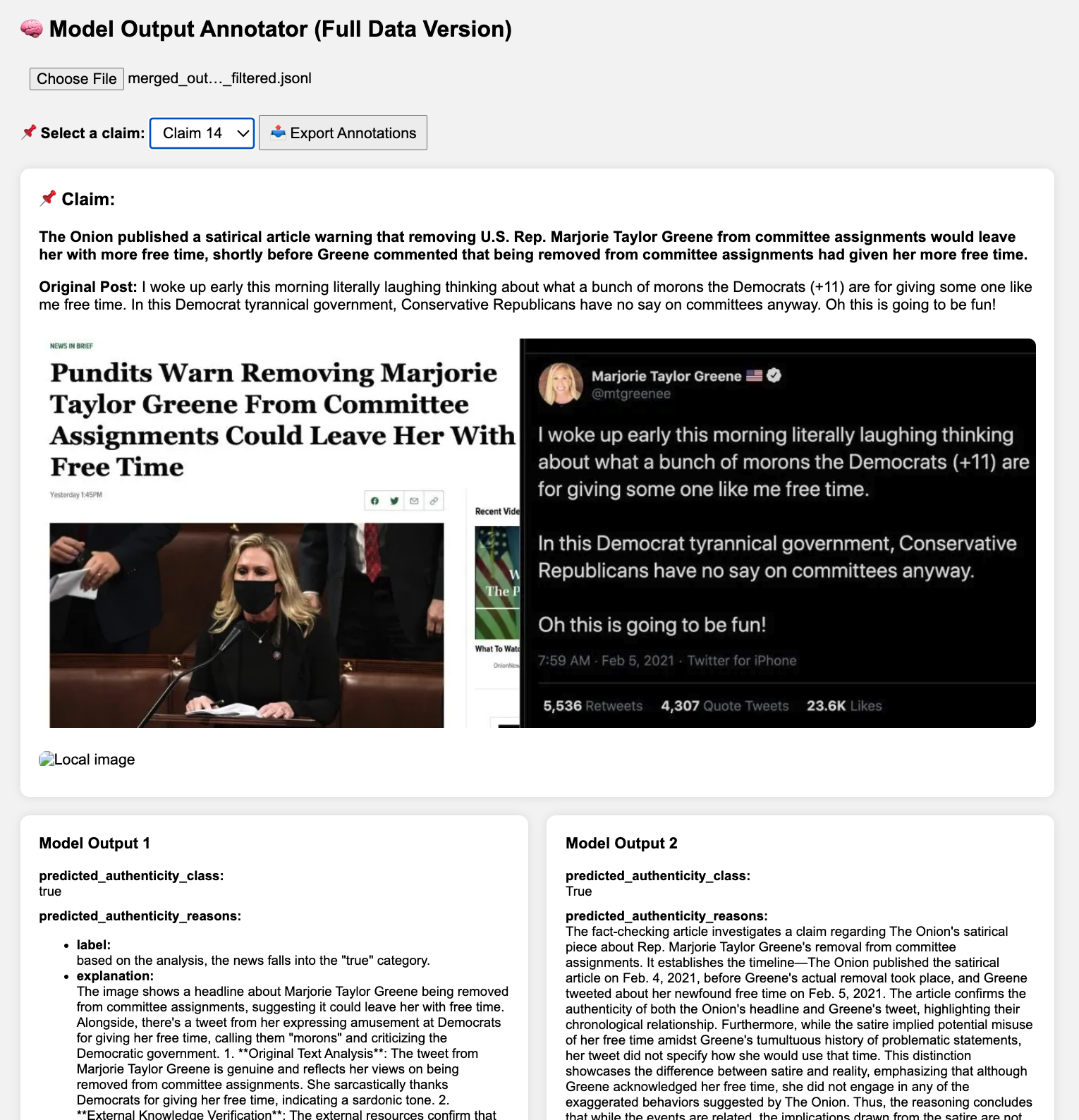}
    \caption{Screenshot of the model output annotator interface.}
    \label{fig:model_annotation_screeshot}
\end{figure}

Forty RW-Post instances are randomly sampled and annotated by three annotators. 
Model outputs and ground-truth references are presented in randomized order with anonymized identifiers to mitigate bias. Figure~\ref{fig:model_annotation_screeshot} shows the interface used by human evaluators to score the outputs of different models, including the ground-truth results.

\textbf{The following guideline was given to human evaluators, outlining the definitions of all evaluation metrics and the corresponding rating criteria.}
\label{human_evaluation_guideine}
\subsubsection*{Task Definition}

The goal of the evaluator (the annotator) is to assess the outputs produced by fact verification models.
Each model aims to determine the authenticity of a given claim and provide corresponding explanations.

Each claim originates from a social media post.
The post text and image are provided as contextual information to assist understanding, but the final evaluation focuses strictly on the claim itself.

\subsubsection*{Evaluation Interface Overview}

In the annotation interface, each claim is presented together with its textual content, image context, and outputs generated by three different models.
Each model output is shown as an independent card containing:
\begin{itemize}
    \item Authenticity judgment
    \item Reasoning
    \item Cited evidence
\end{itemize}

After finishing the evaluation for one claim, the results should be exported as a JSON file named according to the claim ID (e.g., \texttt{1.json}, \texttt{2.json}).
Each model output is presented as an independent card containing:
\begin{tcolorbox}[colback=gray!5, colframe=gray!40, boxrule=0.4pt, arc=2pt]
\ttfamily\small
  "predicted\_authenticity\_class": "...", \\
  "predicted\_authenticity\_reasons": "...", \\
  "predicted\_authenticity\_key\_points": [...], \\
  "evidence\_list": [...], \\
  "image\_analysis\_result": [...], \\
  "confidence\_level": ""
\end{tcolorbox}

\subsubsection*{Reasoning Hallucination}

\paragraph{What to Evaluate}
This dimension evaluates the factual soundness of the model's reasoning.
Specifically, check whether the reasoning provided in \texttt{predicted\_authenticity\_reasons} is directly supported by the evidence listed in \texttt{evidence\_list}.

\paragraph{How to Judge}
\begin{enumerate}
    \item Read the reasoning and key points.
    \item Identify factual claims such as numbers, dates, or causal statements.
    \item Verify whether these claims are supported by the cited evidence.
    \item If no concrete facts are invoked, evaluate logical coherence only.
\end{enumerate}

\paragraph{Target Fields}
\begin{itemize}
    \item \texttt{predicted\_authenticity\_reasons}
    \item \texttt{predicted\_authenticity\_key\_points}
\end{itemize}
Each criterion is rated on a three-level ordinal scale (0--2), mapped from \{\textit{none}, \textit{mild}, \textit{severe}\} to \{0,1,2\} (Table~\ref{tab:reasoning_hallucination}).
\begin{table}[H]
\centering
\small
\begin{tabular}{p{1cm} p{6.5cm}}
% \begin{tabular}{cc}
\hline
\textbf{Score} & \textbf{Criteria} \\
\hline
none   & Reasoning is coherent and well-supported by evidence. \\
mild   & Minor unsupported assumptions without affecting the main conclusion. \\
severe & Core reasoning is speculative or disconnected from evidence. \\
\hline
\end{tabular}

\caption{Scoring criteria for reasoning hallucination.}
\label{tab:reasoning_hallucination}
\end{table}

\subsubsection*{Evidence Usage Hallucination}

\paragraph{What to Evaluate}
This dimension assesses whether the cited evidence is correctly used and accurately represented.

\paragraph{How to Judge}
\begin{enumerate}
    \item Examine each item in \texttt{evidence\_list} and \texttt{image\_analysis\_result}.
    \item Check whether the evidence supports the claims made.
    \item Identify any fabrication, misinterpretation, or irrelevance.
\end{enumerate}

\paragraph{Target Fields}
\begin{itemize}
    \item \texttt{evidence\_list}
    \item \texttt{image\_analysis\_result}
\end{itemize}
Each criterion is rated on a three-level ordinal scale (0--2), mapped from \{\textit{none}, \textit{mild}, \textit{severe}\} to \{0,1,2\} (Table~\ref{tab:evidence_hallucination}).
\begin{table}[H]
\centering
\small
\begin{tabular}{p{1cm} p{6.5cm}}
\hline
\textbf{Score} & \textbf{Criteria} \\
\hline
none    & Evidence is accurate, relevant, and supports the reasoning. \\
mild & Evidence is partially misused or weakly relevant. \\
severe    & Evidence is fabricated or contradicts the conclusion. \\
\hline
\end{tabular}
\caption{Scoring criteria for evidence usage hallucination.}
\label{tab:evidence_hallucination}
\end{table}

\subsubsection*{Label Justification}

\paragraph{What to Evaluate}
This dimension examines whether the predicted authenticity label is justified.

\paragraph{How to Judge}
\begin{enumerate}
    \item Identify the predicted label.
    \item Compare it with the reasoning and evidence.
    \item Check for overconfidence or contradiction.
\end{enumerate}

\paragraph{Target Fields}
\begin{itemize}
    \item \texttt{predicted\_authenticity\_class}
    \item \texttt{predicted\_authenticity\_reasons}
\end{itemize}

\begin{table}[H]
\centering
\small
\begin{tabular}{p{1.5cm} p{6cm}}
\hline
\textbf{Score} & \textbf{Criteria} \\
\hline
justified     & Label is consistent with reasoning and evidence. \\
overconfident & Label is stronger than supported by evidence. \\
hallucinated  & Label contradicts the reasoning or evidence. \\
\hline
\end{tabular}
\caption{Scoring criteria for label justification.}
\label{tab:label_justification}
\end{table}
\clearpage

\end{document}